\documentclass[10pt,twocolumn,letterpaper]{article}

\usepackage[pagenumbers]{iccv} 

\definecolor{iccvblue}{rgb}{0.21,0.49,0.74}
\usepackage[pagebackref,breaklinks,colorlinks,allcolors=iccvblue]{hyperref}

\usepackage{pgfplots}
\usepackage{adjustbox}
\usepackage{diagbox}
\usepackage{gensymb}
\usepackage{array}
\usepackage{xcolor}
\usepackage{pifont}
\newcolumntype{H}{>{\setbox0=\hbox\bgroup}c<{\egroup}@{}}

\usepackage{graphicx}	
\usepackage{amsmath}	
\usepackage{amssymb}	
\usepackage{booktabs}
\usepackage{times}
\usepackage{microtype}
\usepackage{epsfig}
\usepackage[table,xcdraw,dvipsnames]{xcolor}
\usepackage{caption}
\usepackage{float}
\usepackage{placeins}
\usepackage{color, colortbl}
\usepackage{stfloats}
\usepackage{enumitem}
\usepackage{tabularx}
\usepackage{xstring}
\usepackage{multirow}
\usepackage{xspace}
\usepackage{url}
\usepackage{subcaption}
\usepackage{xcolor}
\usepackage[hang,flushmargin]{footmisc}

\definecolor{tablebest}{HTML}{7CB696}
\definecolor{tablesecond}{HTML}{D3E6D1}
\definecolor{tablegain}{HTML}{375B70}
\definecolor{tableloss}{HTML}{844E53}
\definecolor{tableheader}{HTML}{EDF2EF}
\newcommand{\tablemetric}[2]{{\setlength{\fboxsep}{0pt}\ifstrequal{#1}{tablebest}{\colorbox{tablebest!40}{\strut\textbf{#2}}}{\underline{#2}}}}
\newcommand{\tablepos}[1]{\textcolor{tablegain}{#1}}
\newcommand{\tableneg}[1]{\textcolor{tableloss}{#1}}

\begin{document}
\raggedbottom
\setlength{\textfloatsep}{10pt plus 2pt minus 2pt}
\title{Understanding and Optimizing Attention-Based Sparse Matching for Diverse Local Features}
\author{Qiang Wang}

\maketitle
\begin{abstract}
We revisit the training of attention-based sparse image matching models to support diverse local features. We first identify one critical design choice that has been previously overlooked, which significantly affects the performance of the LightGlue. We then investigate the role of detectors and descriptors within the transformer-based matching framework, revealing that detectors, rather than descriptors, are the primary cause for performance difference. Finally, we propose a novel approach to fine-tune existing image matching models using keypoints from a diverse set of detectors, yielding
a universal, detector-agnostic model. When deployed zero-shot on novel detectors, the resulting model achieves or exceeds the accuracy of models specifically trained for those features. Our findings offer practical insights for deploying transformer-based matchers across arbitrary detectors and inform the design of next-generation local features.
\end{abstract}
\section{Introduction}
\label{sec:intro}
Image matching or wide baseline stereo is important for computer vision tasks such as structure from motion~\cite{schonberger2016structure,wu2013towards}, visual localization~\cite{sattler2018benchmarking,sarlin2022lamar,taira2018inloc,panek2025guide,blum2025crocodl} and simultaneous localization and mapping (SLAM)~\cite{mur2017orb}. Many methods have been proposed, among which sparse image matching methods are known for efficiency and robustness~\cite{jin2021image,loiseau2025rubik}.

Sparse image matching methods mostly follow the detect-and-describe pipeline. Classic detector-based methods were dominated by hand-crafted features such as SIFT~\cite{lowe2004distinctive,arandjelovic2012three}, SURF~\cite{bay2006surf} and ORB~\cite{rublee2011orb}. 
Deep learning methods are later introduced to replace hand-crafted heuristics by learning features such as SuperPoint~\cite{detone2018superpoint}, R2D2~\cite{revaud2019r2d2}, DISK~\cite{tyszkiewicz2020disk}, Dedode~\cite{edstedt2024dedode,edstedt2024dedodev2}, SiLK~\cite{gleize2023silk} and XFeat~\cite{potje2024xfeat}. 
Both classic and learned features are initially matched by identifying nearest neighbors or via ratio test~\cite{lowe2004distinctive}. Due to the presence of potential outliers, robust matching techniques are typically employed to filter out suspicious matches~\cite{jin2021image,barath2021gcransac,fischler1981randomransac}.

\begin{figure}[t]
\centering
\includegraphics[width=\linewidth]{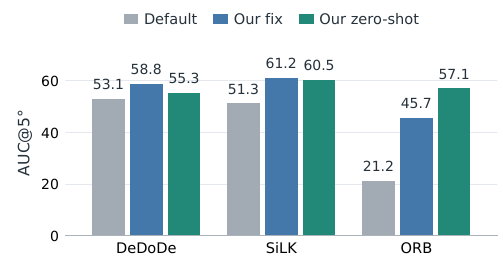}
\caption{\textbf{Matching across local features.} AUC@5\degree on MegaDepth-1500 for the default specialist, our corrected specialist training, and our zero-shot matcher.}
\label{fig:pose-estimation}
\end{figure}

The introduction of transformer-based methods has revolutionized the landscape of image matching. ~\citet{sarlin2020superglue} proposed an attention-based context aggregation method to generate enhanced features and brought significant performance improvement for learned~\cite{detone2018superpoint,tyszkiewicz2020disk,Zhao2023ALIKED} features as well as handcrafted ones~\cite{lowe2004distinctive}. It has been extended to detector-free matching pipelines~\cite{sun2021loftr,chen2022aspanformer,wang2022matchformer}.

However, the training of the SuperGlue model was challenging. Many follow-up works~\cite{viniavskyi2022openglue,shi2022clustergnn,chen2021learning} failed to match the accuracy of the official model. 
LightGlue (LG)~\cite{lindenberger2023lightglue} demonstrated that pre-training on synthetic homographies is crucial and provided a detailed training recipe. 
With glue-factories~\cite{gluefactories}, training LightGlue models became more accessible, but LightGlue models for many detectors such as SiLK and Dedode didn't work well compared to SuperPoint or ALIKED\cite{Zhao2023ALIKED}. 
For example, Kornia~\cite{eriba2019kornia} trained the LightGlue model for Dedode~\cite{edstedt2024dedode}. Its performance is much worse than the model for SuperPoint on MegaDepth-1500 dataset (53.1 compared to 67.0 under AUC@5$^\circ$), as shown in \Cref{fig:pose-estimation}. We find similar results with SiLK\cite{gleize2023silk}. 
Additionally, no prior work has successfully applied attention-based matchers to binary features such as ORB. The underlying reasons for this failure remain less explored.

Upon revisiting the training of existing matching models, we identified that the presence of nearby keypoints—introduced by multi-scale extraction or the absence of non-maximum suppression—can significantly degrade the matching performance. 
Our analysis not only facilitates the training of LightGlue models for binary features such as ORB but also isolates the individual impact of detectors and descriptors on the matching performance. 
For non-binary features, the matching differences of transformer-based matching models are mostly caused by detectors. For binary features, the descriptors contribute more significantly to the final performance.

It is known that the SuperGlue/LightGlue models are feature-dependent~\cite{sarlin2020superglue}. The model trained for SuperPoint features will not work well when applied to other descriptors 
\cite{dukai-blog}. 
Consequently, for each new feature, it's required to train a specialized LightGlue model from scratch. 
Attributing the generalization limitations to the inherent biases of descriptors,~\citet{oblivious} propose a detector-oblivious network, in which existing local descriptors are discarded. They propose a description network to extract local embedding from keypoint location and image, which is jointly trained with the matching model. 
Their model trained with SuperPoint keypoints can work for SIFT detectors, but the performance when applied to other features is unclear.

We show that existing descriptors can work well across different detectors under the attention-based matching framework, as long as the matching model is trained with such detectors. 
Our methods are closely related to the recent works on descriptor compression for visual localization, which show very compact descriptor is only needed\cite{laskar2024dpq}. 
Our analysis is based on decoupling the detector and descriptor for feature matching~\cite{edstedt2024dedode,wang2024mad}. 
We carefully revisit the baseline of applying existing LightGlue models to novel detectors by extracting the descriptors at keypoint locations obtained by the new detector. 
We find such a baseline did not work out-of-the-box, also due to nearby keypoints\cite{oblivious}.  
It can outperform the learned descriptor network~\cite{oblivious} with simple modifications such as extracting at a single scale~\cite{revaud2019r2d2,rublee2011orb} or enabling non-max-suppression (NMS)~\cite{edstedt2024dedode,gleize2023silk}. 

We further propose a method to fine-tune existing image matching models towards universal detector-agnostic ones with fixed descriptors. It's achieved by incorporating additional correspondence supervision from various local detectors during the fine-tuning process. The fine-tuned model demonstrates improved generalization performance when applied to novel detectors. It shows similar performance compared to carefully-tuned specialist matching models. For local detectors without descriptors such as Key.Net~\cite{barroso2019key}, TILDE~\cite{verdie2015tilde}, Dad~\cite{edstedt2025dad}, RaCo\cite{shenoi2026raco}, our model can serve as a drop-in matching network. Considering the detectors rather than descriptors matter more for transformer-based methods, ensembling multiple detectors can boost the matching accuracy. Our model naturally enables efficient detector ensembling since a single matching model is needed during inference.

Our decoupled paradigm also enables powerful matching performance for binary features. Our zero-shot matcher can run successfully on the challenging IMC 2021 dataset with ORB keypoints. It can localize 71.4\% night queries under (0.25m/2\degree) with ORB keypoints and the pretrained LightGlue model on Aachen day-night dataset~\cite{sattler2018benchmarking}. This opens up new applications where the same keypoints need to be computed on-device and in the cloud. For example, replacing the global localization or loop closure in ORB-SLAM3 with transformer-based matching enables tightly-coupled visual positioning systems~\cite{mur2017orb,orbslam3,tightviozju}.

To summarize, the main contributions of the paper are:
\begin{itemize}
\item We revisit the deployment of learning-based sparse image matching models for diverse local features and emphasize the critical importance of removing nearby keypoints.  
\item The impact of detectors and descriptors within attention-based matching frameworks can be decoupled, revealing their respective roles in overall matching performance. 
\item We fine-tune existing models to achieve generic, zero-shot, detector-agnostic matching, demonstrating consistent improvements across all evaluated features.  
\item Our findings offer new insights into optimizing the training and deployment of specialized LightGlue models for novel feature types.
\end{itemize}
\section{Related work}
\label{sec:related}

\textbf{Sparse image matching methods} first detect local features using hand-designed or learned heuristics, represented as 2D locations~\cite{lowe2004distinctive, rublee2011orb, bay2006surf}. Each local feature is then characterized by the descriptor computed from the neighborhood of the keypoint. 
Deep learning methods have been employed to learn local detectors~\cite{barroso2019key, verdie2015tilde}, descriptors~\cite{mishchuk2017working}, or both~\cite{detone2018superpoint, revaud2019r2d2, dusmanu2019d2, edstedt2024dedode}. Over the past decade, the matching accuracy of learned features has surpassed that of traditional methods~\cite{potje2024xfeat, gleize2023silk, edstedt2024dedode}. However, most sparse features still adhere to the detect-and-describe methodology.

Image correspondences are established by nearest neighbor (NN) or ratio test (RT) between the descriptors. 
Nearest neighbor-based matching methods face the issue of false positives and missed matches, as they rely solely on feature appearance without considering the 2D locations of the keypoints~\cite{fischler1981randomransac, chum2005porsac}. Outlier rejection or filtering is often applied to remove suspicious matches~\cite{cavalli2020handcrafted, barath2021gcransac, brachmann2019ngransac, zhao2019nm}.

\noindent\textbf{Deep image matchers} are primarily inspired by the pioneering work of SuperGlue~\cite{sarlin2020superglue, zhang2019oanet, yi2018learning}. The SuperGlue model aggregates both appearance and geometric information with position encoding. Information between pairs of images is shared through cross-attention layers~\cite{vaswani2017attention}. This approach consistently improves performance across various features~\cite{tyszkiewicz2020disk, Zhao2023ALIKED, lowe2004distinctive, eriba2019kornia, bonilla2024mismatchedevaluatinglimitsimage}. Subsequent works aim to enhance SuperGlue by incorporating cluster information~\cite{shi2022clustergnn}, semantic information~\cite{zhang2024mesa} or leveraging foundation models~\cite{jiang2024omniglue, oquab2023dinov2}. Recent studies show that the SuperGlue model still demonstrates superior performance on various tasks~\cite{bonilla2024mismatchedevaluatinglimitsimage}. 
LightGlue~\cite{lindenberger2023lightglue} proposes adaptive keypoint pruning and early exit to speed up the SuperGlue model. It also replaces MLP positional encoding with rotary position encoding~\cite{su2024roformer} and discards the Sinkhorn algorithm for match assignment. The open-sourced glue-factory~\cite{gluefactories} enables easier training of LightGlue models for novel features.

\begin{table}[t]
\centering
\footnotesize
\setlength{\tabcolsep}{3pt}
\renewcommand{\arraystretch}{1.18}
\begin{adjustbox}{max width=\linewidth}
\begin{tabular}{@{}l l cc cc@{}}
\toprule
\multirow{2}{*}{\textbf{Detector}} & \multirow{2}{*}{\textbf{Setting}} &
\multicolumn{2}{c}{Default} & \multicolumn{2}{c}{\cellcolor{tableheader}\textbf{Ours}} \\
\cmidrule(lr){3-4}\cmidrule(l){5-6}
& & AUC $\uparrow$ & \#NI $\uparrow$ & AUC $\uparrow$ & \#NI $\uparrow$ \\
\midrule
\multirow{2}{*}{DeDoDe} & NMS $r=3$ & 54.7/69.8/81.2 & \tablemetric{tablebest}{465} & \tablemetric{tablebest}{58.8}/\tablemetric{tablebest}{73.1}/\tablemetric{tablebest}{83.3} & \tablemetric{tablesecond}{458} \\
 & no NMS & 53.1/66.6/77.9 & 275 & \tablemetric{tablesecond}{56.2}/\tablemetric{tablesecond}{70.5}/\tablemetric{tablesecond}{81.7} & 254 \\
\midrule
\multirow{2}{*}{SiLK} & NMS $r=3$ & \tablemetric{tablesecond}{57.9}/\tablemetric{tablesecond}{71.9}/\tablemetric{tablesecond}{82.2} & \tablemetric{tablesecond}{329} & \tablemetric{tablebest}{61.2}/\tablemetric{tablebest}{74.8}/\tablemetric{tablebest}{84.4} & \tablemetric{tablebest}{348} \\
 & no NMS & 51.3/65.0/75.8 & 131 & 47.8/61.8/72.7 & 94 \\
\midrule
\multirow{2}{*}{R2D2} & SS & 58.4/70.7/79.3 & \tablemetric{tablesecond}{571} & \tablemetric{tablebest}{64.1}/\tablemetric{tablesecond}{76.8}/\tablemetric{tablesecond}{85.7} & \tablemetric{tablebest}{579} \\
 & MS & \tablemetric{tablesecond}{63.9}/\tablemetric{tablebest}{77.3}/\tablemetric{tablebest}{86.5} & 329 & 60.1/74.1/83.7 & 434 \\
\midrule
\multirow{2}{*}{ORB} & SS & \textemdash & \textemdash & \tablemetric{tablebest}{45.7}/\tablemetric{tablebest}{59.4}/\tablemetric{tablebest}{71.1} & \tablemetric{tablebest}{173} \\
 & MS & \textemdash & \textemdash & \tablemetric{tablesecond}{21.2}/\tablemetric{tablesecond}{32.6}/\tablemetric{tablesecond}{45.4} & \tablemetric{tablesecond}{94} \\
\bottomrule
\end{tabular}
\end{adjustbox}
\caption{\textbf{Effect of keypoint extraction.} Pose AUC@$5^\circ/10^\circ/20^\circ$ and inlier count (\#NI) on MegaDepth-1500, with at most 2048 features and LO-RANSAC. Green/bold: best; underline: second-best, per detector and metric. SS/MS: single-/multi-scale.}
\label{table_dedode_ft}
\vspace{-0.05in}
\end{table}

\noindent\textbf{Decoupling the feature detector and descriptor} is not novel. It has been explored for both handcrafted features~\cite{harris1988combined, shi1994good} and learned features~\cite{barroso2019key, mishchuk2017working, mukundan2019mkd, tian2019sosnet}. While some works emphasize the advantages of joint learning~\cite{detone2018superpoint, dusmanu2019d2},~\citet{edstedt2024dedode} argue that decoupled learning is superior. However, the decoupling of the detector and descriptor is rarely addressed when using attention-based transformer models~\cite{oblivious}, partially due to the poor generalization of SuperGlue models to other features~\cite{viniavskyi2022openglue,shi2022clustergnn,chen2021learning}.

\noindent\textbf{Generalization of deep image matching models} to new features has rarely been discussed since it was commonly believed that the SuperGlue models are feature-dependent. The SuperGlue model trained for SuperPoint will fail when fed with R2D2 descriptors~\cite{revaud2019r2d2,oblivious}. This is expected since the domain gap between different descriptors can be large. 
As an early attempt to build a detector-agnostic image matching model,~\citet{oblivious} propose the detector-oblivious network. 
A detector-oblivious description network is fed with the keypoint locations and image to replace local descriptors. 
Some preliminary results are shown, which successfully apply such a model trained on SuperPoint keypoints to SIFT keypoints. 
They also include a simple baseline, which naively extracts the SuperPoint descriptor from the R2D2 keypoints and feeds it into the SuperGlue model trained for SuperPoint features. However, such a baseline causes significant performance drops on the MegaDepth dataset. We revisit this baseline with detailed analysis and find that several implementation details can significantly impact performance. By addressing these issues, we obtain a strong zero-shot image matching model for novel detectors without modifying the network architecture or retraining from scratch.

\label{sec:method}
\begin{figure}[t!]
\centering
\includegraphics[width=\linewidth]{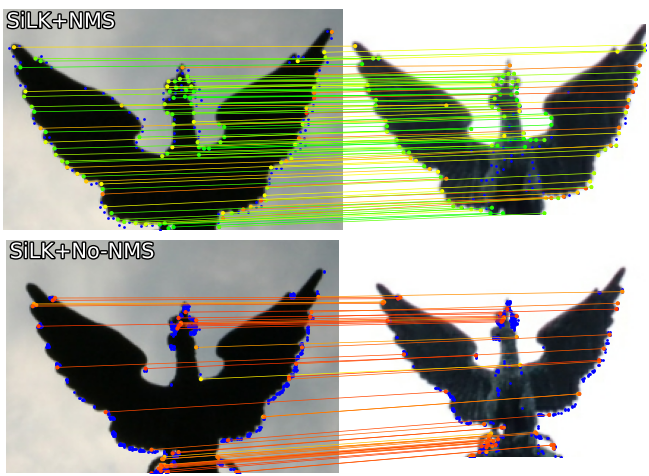}
\caption{\textbf{Nearby keypoints and matching.} LightGlue matching with SiLK~\cite{gleize2023silk}, with NMS (top) and without NMS (bottom). Closely spaced keypoints yield many unmatched correspondences.}
\label{fig:SiLK-nms}
\vspace{-0.05in}

\end{figure}

\section{Always remove nearby keypoints}
During training of the SuperGlue models, ground truth correspondences are generated by first computing the reprojection matrix between all detected keypoints using the ground truth homography or pose and depth. Correspondences are cells with a reprojection error that is a minimum along both rows and columns, and that is lower than the given threshold: 3, 5, and 3 pixels for homographies, indoor, and outdoor matching respectively~\cite{sarlin2020superglue}. For SuperPoint or DISK, which perform detection at a single scale and apply non-maximum suppression, SuperGlue works effectively~\cite{lindenberger2023lightglue}. 

For features that might generate nearby keypoints at adjacent locations as shown in \cref{fig:SiLK-nms}, establishing ground truth matches using the criteria above can be problematic. On one hand, R2D2 and ORB extract keypoints at multiple scales, which may produce keypoints at very close locations on different scales. Currently the correspondences for features across different scales are not handled explicitly.

On the other hand, recent local features such as DeDoDe~\cite{edstedt2024dedodev2} and SiLK~\cite{gleize2023silk} show strong matching results with MNN at the cost of many keypoints (10K-30K), but
the official implementations of DeDoDe and SiLK come without non-maximum suppression (NMS)~\cite{edstedt2024dedodev2}. When integrating DeDoDe with Kornia~\cite{eriba2019kornia}, the code related to NMS has been completely removed. Due to the inaccurate depth estimation on MegaDepth dataset, correspondences between nearby keypoints can become unreliable or incorrect. 

To verify the impact of NMS and multi-scale extraction, we train four LightGlue models: DeDoDe and SiLK features with NMS enabled, R2D2/ORB with single-scale extraction. 
The results for two-view pose estimation task on MegaDepth-1500 dataset are shown in \cref{table_dedode_ft}. 

Our model for DeDoDe trained with NMS outperforms the official model provided by Kornia. With NMS at inference, AUC@5$^\circ$ improves from 54.7 to 58.8 (+4.1 points); without NMS, it improves from 53.1 to 56.2 (+3.1 points). The same trend holds for SiLK features as well, with NMS during training and testing, we obtain 9.9\% absolute gain. For R2D2, we obtain similar accuracy when the training and inference configurations are aligned. Notable performance drops (4\%--6.1\%) occur when the inference configuration differs from the training setup for R2D2, indicating an inherent bias in the matching models.

We visualize the matching results of the LightGlue model on SiLK features in \cref{fig:SiLK-nms}, with and without NMS. Without NMS, the SiLK features contain many repeatable but closely spaced keypoints. However, the LightGlue model fails to match many of these potential correspondences. When NMS is applied, the same LightGlue model successfully recovers most of the matches. Interestingly, this behavior contrasts with the original SiLK paper, which recommends disabling NMS when using nearest-neighbor matching.

Training the default LightGlue model on multi-scale ORB features fails to converge. Switching to single-scale extraction stabilizes training. For LightGlue with ORB features, the AUC@5$^\circ$ is 45.7, while nearest-neighbor matcher obtains only 10.1, suggesting LightGlue can bring significant boost for binary features since it can leverage the context to mitigate the deficiencies of less discriminative descriptors. Compared to non-binary features, LightGlue models still perform worse. In the following section, we identify the individual contributions of detector and descriptor for the performance gap.

\section{Generalization of attention-based matcher}
\subsection{Baseline setup and results}

\begin{figure}[t]
	\centering
	\includegraphics[width=\linewidth]{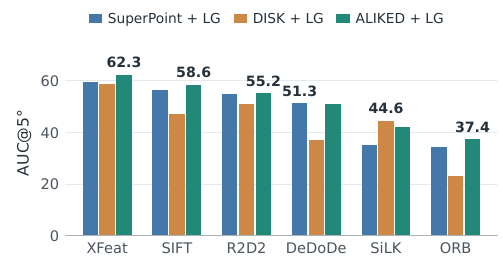}
	\caption{\textbf{Cross-detector generalization.} Off-the-shelf LightGlue models on MegaDepth-1500. Colors identify the descriptor--matcher pair; labels mark the highest AUC@5\degree for each detector.}
        \label{fig-gene}
\end{figure} 

We study the generalization of the attention-based matcher with a simple baseline that applies existing LightGlue model to novel features. Given a LightGlue model trained for feature $F_A$ and any feature detector $F_B$ that outputs the keypoint position $P$, we sample the corresponding descriptor from the feature map of $F_A$. We feed location $P$ and feature descriptions (denoted as $F_{A,B}$) to the pre-trained LightGlue model trained for $F_A$. 

We run the baseline on MegaDepth-1500~\cite{li2018megadepth} dataset for two-view relative pose estimation task. 
The pose accuracy measured with AUC@ 5$^\circ$ /10$^\circ$ is shown in \cref{fig-gene}. Different detectors exhibit varying performances when matched with off-the-shelf models, indicating such baseline doesn't always work out-of-the-box.

XFeat detector~\cite{potje2024xfeat} is designed to be a faster alternative to DISK/SuperPoint. It can be matched reasonably well with off-the-shelf ALIKED matcher (62.3/76.0, AUC@5$^\circ$/10$^\circ$). For reference, the accuracy of the XFeat features matched with nearest neighbor matching is only 42.8/57.1. 
SIFT detector with the specialist LightGlue model for SIFT descriptors achieves accuracy of 59.2 (AUC@5$^\circ$)~\cite{lindenberger2023lightglue}. When replacing the SIFT matcher with SuperPoint or ALIKED matchers, the respective results are 56.3 ($\downarrow$2.9) and 58.6 ($\downarrow$0.6). 
These results are encouraging, considering that SuperPoint primarily focuses on corners, while SIFT is designed to detect blobs and all model weights are frozen.

\begin{figure}[t]
	\centering
	\includegraphics[width=\linewidth]{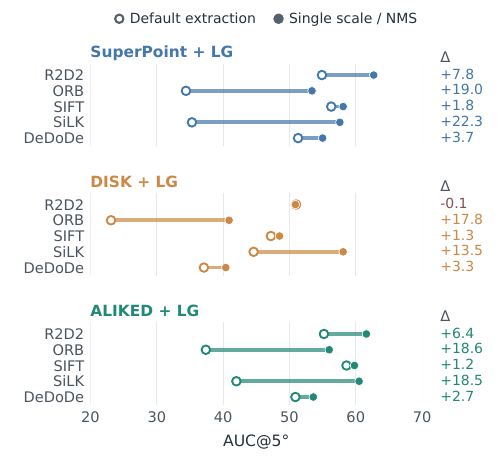}
	\caption{\textbf{Effect of keypoint extraction.} Open/filled circles show default/modified extraction on MegaDepth-1500. R2D2, ORB, and SIFT use single-scale extraction; SiLK and DeDoDe use NMS ($r=3$). $\Delta$ is the signed AUC@5\degree change in points.}
    \label{fig-ft}
\end{figure} 

DeDoDe, SiLK, and ORB exhibit poor results when paired with existing matchers, consistent with previous works~\cite{oblivious}.  As we have discussed before, the off-the-shelf LightGlue models are trained without nearby keypoints, we therefore extract features at a single scale for R2D2, SIFT, and ORB. The results are shown in \cref{fig-ft}. This leads to significant improvements. When matched with SuperPoint descriptors as in~\cite{oblivious}, extracting single-level features for R2D2 results in 14.2\%  relative gain  (54.9$\rightarrow$62.7). The number of inliers also increases substantially (255$\rightarrow$606). Similar results are obtained for the DISK and ALIKED matchers. 
SIFT features consistently benefit from extracting at single-scale only, though the impact of multi-scale extraction on SIFT is relatively low compared to ORB and R2D2 ($\uparrow$1.2\%--1.8\% @AUC/5$^\circ$ with LO-RANSAC). The reason that SIFT is the least affected (2.4\%) is that keypoints are detected by identifying local extrema in the difference of Gaussian thus making it unlikely to obtain multiple keypoints at nearby locations. 
ORB detects feature independently across many scales and the locations with high responses are likely to be identified as keypoints multiple times across different scales. This causes a major difference between single-scale and multi-scale extraction (117\%).

Out of the three matching models evaluated, the LightGlue model for ALIKED generalizes best. 
When coupled with SIFT keypoints, it obtains comparable results (58.6/72.8, AUC@5$^\circ$/10$^\circ$) with the official LightGlue model trained for SIFT features (59.2/72.6,AUC@5$^\circ$/10$^\circ$). 
For DeDoDe keypoints, its performance (AUC@5$^\circ$ 53.6, inliers 544) is comparable to 
the specialist LightGlue model for DeDoDe (AUC@5$^\circ$ 54.7, inliers 465). The LightGlue model for SuperPoint descriptor shows competitive results with slightly worse accuracy. DISK descriptors and matcher, on the other hand, perform worst for most detectors. This indicates that the DISK descriptor–matcher pair may exhibit stronger bias, limiting its generalization.

Given that the descriptor model and LightGlue models are specialists trained for specific detectors, the domain gap when applying them to different detectors may lead to unexpected performance drops. 
In the following section, we present a method to fine-tune existing matching models towards stronger detector-agnostic models. It not only boosts performance for all off-the-shelf models, but also closes the gap between them.

\subsection{Towards detector-agnostic matching model}
\begin{figure}[t]
\centering
\includegraphics[width=\linewidth]{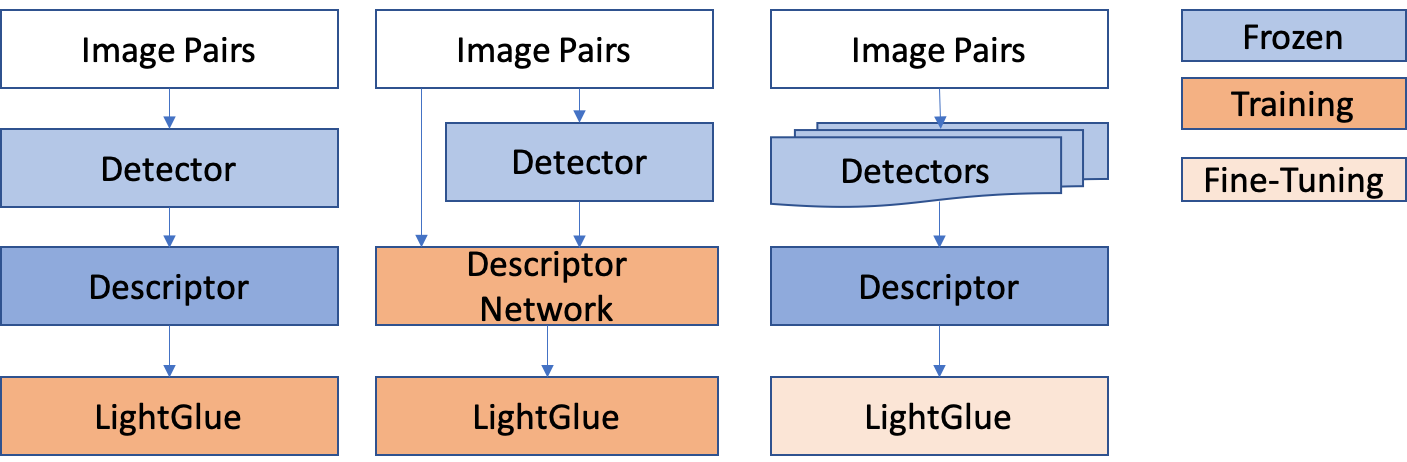}
\caption{Left: Standard LightGlue training for a single detector.
Middle: Detector-agnostic training by jointly learning descriptors and LightGlue~\cite{oblivious}.
Right: Our fine-tuning of an existing LightGlue model to improve cross-detector generalization.}
\label{fig:framework}
\end{figure}


Previous works~\cite{oblivious} attribute the representation power of the descriptor as the bottleneck of generalization and replace the existing descriptor with a learned network. We argue that most modern descriptors are sufficiently discriminative for transformer-based matchers, which can leverage cross-attention and contextual information effectively. We attribute the performance gap with the above model to the bias of local detectors during training rather than descriptors. The keypoints for training are all extracted using one specific detector. 
To verify the idea that existing descriptors are not the issue, 
we fine-tune the LightGlue model with descriptors extracted by multiple detectors, while keeping the descriptor network weights frozen, as shown in \cref{fig:framework}.

Three LightGlue models~\cite{lindenberger2023lightglue} for SuperPoint, DISK, and ALIKED  are tuned on the MegaDepth dataset~\cite{li2018megadepth}, which is also used for initial training. We extract keypoints for all three features and sample the same descriptors for all keypoints. The results indicate that the detectors, not the descriptors, are the key factor that determines the accuracy of transformer-based matching methods.

In the following section, we evaluate the fine-tuned model on detectors that are not seen during training or fine-tuning. Our fine-tuned model can be applied out-of-the-box to detectors that produce keypoints but lack associated descriptors (e.g., DaD~\cite{edstedt2025dad} or RaCo~\cite{shenoi2026raco}), without requiring any additional training.

\section{Experiments}

\label{sec:result}
We evaluate the proposed model on the relative pose estimation task on the MegaDepth-1500 dataset and Air-to-Ground dataset~\cite{chen2025rdd}, followed by the evaluation on the Image Matching Competition 2021 (IMC) dataset. For downstream tasks, we evaluate the performance on the Aachen Day-Night dataset~\cite{sattler2018benchmarking} and the InLoc~\cite{taira2018inloc} dataset for visual localization tasks.

\subsection{Relative pose estimation}
For the relative pose estimation task, the MegaDepth-1500 dataset is used following previous work~\cite{sun2021loftr,lindenberger2023lightglue,potje2024xfeat}. The dataset contains 1500 image pairs captured in St. Peter's Square \& Reichstag.

\noindent\textbf{Evaluation setup}: The long side of the image is resized to 1600 and at most 2048 keypoints are extracted for all detectors. The essential matrix is computed with LO-RANSAC~\cite{PoseLib}. Rotation and translation are then computed following~\cite{lindenberger2023lightglue}.
The inlier thresholds are selected on a separate validation set.
We use the official DeDoDe (v1) weights since it is the version used by Kornia to train the LightGlue model. 
The pose error is computed as the maximum angular error in rotation and translation. We report AUC at 5$^\circ$, 10$^\circ$, and 20$^\circ$.

\noindent\textbf{Baselines:} We use nearest neighbor (NN) and LightGlue models trained individually for each feature (specialist LightGlue). For SIFT, the official weights provided by LightGlue~\cite{lindenberger2023lightglue} are used. For DeDoDe, the LightGlue model provided by Kornia~\cite{eriba2019kornia} is used. For R2D2, XFeat and ORB, we train with glue-factory\cite{gluefactories}. For R2D2 and ORB, we extract only at a single scale during training and inference.

We include GIM~\cite{xuelun2024gim}, which is an optimized SuperGlue model trained on both the MegaDepth~\cite{li2018megadepth} dataset and additional 100 hours of internet videos. 
We include a re-implemented version of detector-oblivious network (DON)~\cite{oblivious}, which jointly learns the descriptor with keypoint location and image. As the specific description network is not disclosed, it is implemented as a 4-layer MLP network. A patch size $P \times P$ centered at the keypoint location is cropped and fed into the MLP network, which outputs a 256-D descriptor as input to the LightGlue model. The weights for MLP are jointly trained with the LightGlue network. This is similar to the patch embedding of Vision Transformer~\cite{Dosovitskiy2020vit}. The auxiliary networks for overlap and depth regions estimation~\cite{oblivious} are removed for fair comparison.

\begin{table*}[t]
\centering
\small
\setlength{\tabcolsep}{4pt}
\renewcommand{\arraystretch}{1.12}
\begin{adjustbox}{max width=\textwidth}
\begin{tabular}{@{}l c c c c c c@{}}
\toprule
\rowcolor{tableheader}
\textbf{Matcher} & \textbf{SIFT} & \textbf{DeDoDe} & \textbf{R2D2} & \textbf{XFeat} & \textbf{ORB} & \textbf{Avg.} \\
\midrule

Nearest Neighbor
&40.1/53.5/64.6
&44.3/58.1/69.0
&45.4/56.3/64.3
&42.8/57.1/68.6
&10.1/16.7/24.7
&36.5/48.3/58.2 \\

DON~\cite{oblivious}
&45.8/58.9/69.3
&45.4/60.1/72.3
&52.7/65.7/75.5
&56.8/71.7/82.3
&48.2/61.4/71.4
&49.8/63.6/74.2 \\

GIM~\cite{xuelun2024gim}
&57.7/71.7/82.0
&\tablemetric{tablebest}{57.6}/\tablemetric{tablebest}{72.7}/\tablemetric{tablebest}{83.5}
&62.4/75.2/84.6
&60.0/74.4/84.3
&54.0/69.0/80.1
&58.3/\tablemetric{tablesecond}{72.6}/82.9 \\

LightGlue~\cite{lindenberger2023lightglue}
&\tablemetric{tablesecond}{59.2}/72.6/82.8
&54.7/69.8/81.2
&\tablemetric{tablebest}{64.1}/\tablemetric{tablebest}{76.8}/\tablemetric{tablesecond}{85.7}
&60.7/74.9/84.8
&45.7/59.4/71.1
&56.9/70.7/81.1 \\

\midrule

DISK+LG
&47.2/59.2/68.0
&40.4/54.3/65.7
&50.9/62.8/71.1
&58.9/72.8/82.7
&40.9/52.0/60.2
&47.7/60.2/69.5 \\

\quad \textbf{+ Ours}
&55.2/68.7/78.3
&51.0/66.2/78.0
&60.1/72.5/81.4
&62.0/75.6/85.2
&55.7/69.0/79.0
&56.8/70.4/80.4 \\

\quad $\Delta$
&{\scriptsize \tablepos{+8.0}/\tablepos{+9.5}/\tablepos{+10.3}}
&{\scriptsize \tablepos{+10.6}/\tablepos{+11.9}/\tablepos{+12.3}}
&{\scriptsize \tablepos{+9.2}/\tablepos{+9.7}/\tablepos{+10.3}}
&{\scriptsize \tablepos{+3.1}/\tablepos{+2.8}/\tablepos{+2.5}}
&{\scriptsize \tablepos{+14.8}/\tablepos{+17.0}/\tablepos{+18.8}}
&{\scriptsize \tablepos{+9.1}/\tablepos{+10.2}/\tablepos{+10.9}} \\

\midrule

SP+LG
&56.3/70.6/81.3
&55.0/69.7/80.9
&62.7/75.4/84.6
&59.6/74.0/84.1
&53.4/67.9/78.8
&57.4/71.5/81.9 \\

\quad \textbf{+ Ours}
&58.0/72.4/\tablemetric{tablesecond}{83.3}
&\tablemetric{tablesecond}{57.1}/\tablemetric{tablesecond}{71.2}/\tablemetric{tablesecond}{82.6}
&63.2/76.0/85.4
&60.8/75.1/85.1
&53.2/68.1/80.0
&\tablemetric{tablesecond}{58.5}/\tablemetric{tablesecond}{72.6}/\tablemetric{tablesecond}{83.3} \\

\quad $\Delta$
&{\scriptsize \tablepos{+1.7}/\tablepos{+1.8}/\tablepos{+2.0}}
&{\scriptsize \tablepos{+2.1}/\tablepos{+1.5}/\tablepos{+1.7}}
&{\scriptsize \tablepos{+0.5}/\tablepos{+0.6}/\tablepos{+0.8}}
&{\scriptsize \tablepos{+1.2}/\tablepos{+1.1}/\tablepos{+1.0}}
&{\scriptsize \tableneg{-0.2}/\tablepos{+0.2}/\tablepos{+1.2}}
&{\scriptsize \tablepos{+1.1}/\tablepos{+1.1}/\tablepos{+1.4}} \\

\midrule

ALIKED+LG
&58.6/\tablemetric{tablesecond}{72.8}/82.9
&53.6/68.9/80.5
&61.6/74.5/83.9
&\tablemetric{tablesecond}{62.3}/\tablemetric{tablesecond}{76.0}/\tablemetric{tablesecond}{85.6}
&\tablemetric{tablesecond}{56.0}/\tablemetric{tablesecond}{70.1}/\tablemetric{tablesecond}{80.7}
&58.4/72.5/82.7 \\

\quad \textbf{+ Ours}
&\tablemetric{tablebest}{60.1}/\tablemetric{tablebest}{74.4}/\tablemetric{tablebest}{84.4}
&55.3/70.6/82.0
&\tablemetric{tablesecond}{63.4}/\tablemetric{tablesecond}{76.3}/\tablemetric{tablebest}{85.8}
&\tablemetric{tablebest}{62.7}/\tablemetric{tablebest}{76.5}/\tablemetric{tablebest}{86.0}
&\tablemetric{tablebest}{57.1}/\tablemetric{tablebest}{71.1}/\tablemetric{tablebest}{81.7}
&\tablemetric{tablebest}{59.7}/\tablemetric{tablebest}{73.8}/\tablemetric{tablebest}{84.0} \\

\quad $\Delta$
&{\scriptsize \tablepos{+1.5}/\tablepos{+1.6}/\tablepos{+1.5}}
&{\scriptsize \tablepos{+1.7}/\tablepos{+1.7}/\tablepos{+1.5}}
&{\scriptsize \tablepos{+1.8}/\tablepos{+1.8}/\tablepos{+1.9}}
&{\scriptsize \tablepos{+0.4}/\tablepos{+0.5}/\tablepos{+0.4}}
&{\scriptsize \tablepos{+1.1}/\tablepos{+1.0}/\tablepos{+1.0}}
&{\scriptsize \tablepos{+1.3}/\tablepos{+1.3}/\tablepos{+1.3}} \\

\bottomrule
\end{tabular}
\end{adjustbox}

\caption{\textbf{Generalization to unseen detectors.} Pose AUC@$5^\circ/10^\circ/20^\circ$ on MegaDepth-1500. Green/bold and underlined values mark the best and second-best results per metric. Avg. is the equal-weight mean over five detectors; $\Delta$ is fine-tuned minus baseline.}
\label{tab:tuned_generalization}

\vspace{-0.05in}
\end{table*}

\noindent\textbf{Results:} 
As shown in \Cref{tab:tuned_generalization}, the LightGlue model trained for each feature outperforms NN-based matching, indicating the effectiveness of attention-based context aggregation models~\cite{sarlin2020superglue,lindenberger2023lightglue}. 
Our fine-tuned model brings consistent improvements for all detectors.  
For SIFT keypoints, our fine-tuned model with ALIKED descriptor outperforms the official LightGlue model for SIFT. 
For DeDoDe keypoints, our model with SuperPoint descriptors outperforms the LightGlue model for DeDoDe. The results suggest that training the feature-specific matcher might be sub-optimal.

With our fine-tuned model for ALIKED descriptors, we obtain the best matching results for SIFT, XFeat and ORB keypoints. It serves as a strong zero-shot detector-agnostic matching model with an average accuracy of 59.7/73.8/84.0.

The official DISK+LightGlue model did not perform well compared to the LightGlue model for ALIKED, with the average accuracy being 47.7/60.2/69.5. After tuning with our method, the performance is significantly enhanced  (56.8/70.4/80.4). This substantiates the importance of our cross-feature fine-tuning.

Our model trained for SuperPoint features achieves similar performance to GIM. GIM performs better than our model for SIFT features. Note that GIM is trained with annotations derived from SIFT matches on additional internet videos. Our model is entirely fine-tuned on the MegaDepth dataset without SIFT keypoints. Our model with SuperPoint descriptors performs better than GIM for R2D2 and XFeat features, while GIM is better for ORB features.

\begin{figure}[t]
\centering
\includegraphics[width=\linewidth]{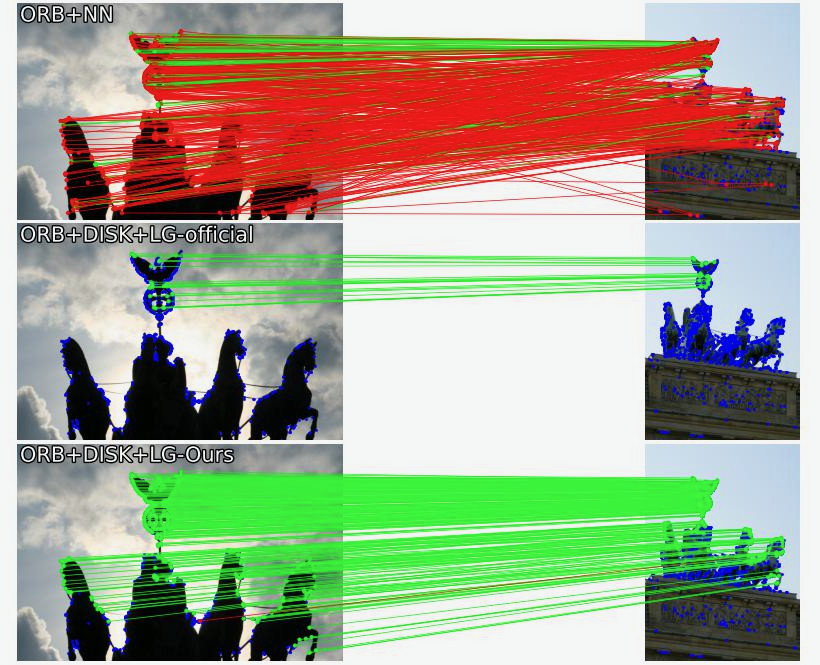}
\caption{\textbf{Matching ORB keypoints.} Top: ORB descriptor + nearest neighbor matching. Middle: DISK descriptor + the official LightGlue model. Bottom: DISK descriptor + our fine-tuned LightGlue model.}
\label{fig:vis_orb}
\vspace{-0.05in}
\end{figure}

Matching ORB features with Hamming distance results in many wrong matches as shown in \Cref{fig:vis_orb}. With the default LightGlue models, only a few correct matches are kept. Our fine-tuned model produces significantly more correct matches. 

The detector-oblivious network (DON)~\cite{oblivious} performs worse than our descriptor-based matcher, suggesting local descriptors are still important for existing detector-based models. We hypothesize that sufficiently large models may not require the inductive bias of handcrafted descriptors, as evidenced by the recent success of DUSt3R\cite{wang2024dust3r} and follow-up works such as MASt3R\cite{leroy2024grounding} and VGGT\cite{wang2025vggt}.

\subsection{Image Matching Challenge 2021}
\noindent\textbf{Evaluation setup}: We benchmark our model on the Image Matching Challenge 2021~\cite{imc2021}. The results are evaluated on the stereo task and the multi-view task with the PhotoTourism subset. We evaluate the restricted keypoints category, where at most 2048 keypoints can be extracted. The initial matches are filtered with DEGENSAC~\cite{Chum2005,Chum2003,Mishkin2015MODS} before evaluation. The results are reported with mean average accuracy (mAA) at a 10$^\circ$ error threshold in \Cref{table:IMC2021}.

\noindent\textbf{Results:} The off-the-shelf LightGlue model trained for DISK fails when coupled with ORB and R2D2 detectors for the multi-view task, while our fine-tuned model produces valid results for all scenes. With DISK+LightGlue, our fine-tuned model shows better results for all other detectors, with negligible degradation for DISK itself.

Our fine-tuned models for SuperPoint and ALIKED features increase the matching accuracy for most detectors. 
When used as a zero-shot matcher for ORB detectors, our model obtains 64.8 mAA @10$^\circ$ for the multi-view task.

Similar to the results on the MegaDepth-1500 dataset, the multi-scale R2D2 (R2D2-MS) performs worse than single-scale R2D2 (R2D2-SS). 
For example, with our fine-tuned DISK matcher, the mAA@10$^\circ$ for R2D2-SS is 50.0/75.2 for the stereo task, while for R2D2-MS the corresponding accuracy is only 45.3/70.8. Similarly, ORB extracted at a single scale (ORB-SS) outperforms multi-scale ORB (ORB-MS) for both tasks.


\subsection{Air-to-Ground benchmark} 
We evaluate the proposed methods on the Air-to-Ground benchmark~\cite{chen2025rdd}. We evaluate several recent detectors, including DaD\cite{edstedt2025dad}, RaCo\cite{shenoi2026raco}, RDD\cite{chen2025rdd} and DeDoDe\cite{edstedt2024dedodev2}.

Our method brings consistent improvements as shown in \Cref{table:air-ground2}. The improvement is particularly significant for DeDoDe-V2 + ALIKED, where our approach increases AUC from 37.7 to 45.0 at $5^{\circ}$ and from 61.6 to 70.8 at $20^{\circ}$. Similar gains are observed for modern detectors. DaD + ALIKED improves from 53.8 to 56.9 at $5^{\circ}$ and from 78.5 to 80.8 at $20^{\circ}$, while RaCo + ALIKED increases from 54.4 to 56.3 at $5^{\circ}$. Consistent improvements are also observed when replacing ALIKED with DeDoDe-G descriptors.
\begin{table}[t]
\centering\footnotesize
\setlength{\tabcolsep}{3.2pt}
\renewcommand{\arraystretch}{1.12}
\begin{adjustbox}{max width=\linewidth}
\begin{tabular}{@{}l cc >{\columncolor{tableheader}}c >{\columncolor{tableheader}}c cc@{}}
\toprule
\multirow{2}{*}{Detector} & \multicolumn{2}{c}{Default} & \multicolumn{2}{c}{\cellcolor{tableheader}\textbf{Ours}} & \multicolumn{2}{c}{$\Delta$} \\
\cmidrule(lr){2-3}\cmidrule(lr){4-5}\cmidrule(l){6-7}
& Stereo & MV & Stereo & MV & Stereo & MV \\
\midrule
\multicolumn{7}{l}{\textit{DISK + LightGlue}} \\
SuperPoint  & 41.3 & 68.9 & 51.7 & 74.3 & \tablepos{+10.4} & \tablepos{+5.4} \\
DISK        & 59.3 & 76.2 & 58.9 & 76.1 & \tableneg{-0.4} & \tableneg{-0.1} \\
ALIKED      & 57.7 & 72.0 & 61.1 & 76.7 & \tablepos{+3.4} & \tablepos{+4.7} \\
R2D2-SS     & 36.6 & \textcolor{gray}{N/A} & 50.0 & 75.2 & \tablepos{+13.4} & \textemdash \\
R2D2-MS     & 36.3 & \textcolor{gray}{N/A} & 45.3 & 70.8 & \tablepos{+9.0} & \textemdash \\
ORB-SS      & 30.5 & \textcolor{gray}{N/A} & 38.7 & 64.5 & \tablepos{+8.2} & \textemdash \\
ORB-MS      & 18.2 & \textcolor{gray}{N/A} & 25.3 & \textcolor{gray}{N/A} & \tablepos{+7.1} & \textemdash \\
\midrule
\multicolumn{7}{l}{\textit{SuperPoint + LightGlue}} \\
SuperPoint  & 50.1 & 74.5 & 53.8 & 75.1 & \tablepos{+3.7} & \tablepos{+0.6} \\
DISK        & 53.0 & 74.9 & 52.9 & 73.8 & \tableneg{-0.1} & \tableneg{-1.1} \\
ALIKED      & 58.0 & 76.3 & 61.0 & 77.8 & \tablepos{+3.0} & \tablepos{+1.5} \\
R2D2-SS     & 56.8 & 76.9 & 57.6 & 77.0 & \tablepos{+0.8} & \tablepos{+0.1} \\
R2D2-MS     & 46.3 & 71.1 & 47.8 & 72.0 & \tablepos{+1.5} & \tablepos{+0.9} \\
ORB-SS      & 31.0 & 59.2 & 32.3 & 59.3 & \tablepos{+1.3} & \tablepos{+0.1} \\
ORB-MS      & 14.6 & \textcolor{gray}{N/A} & 15.7 & \textcolor{gray}{N/A} & \tablepos{+1.1} & \textemdash \\
\midrule
\multicolumn{7}{l}{\textit{ALIKED + LightGlue}} \\
SuperPoint  & 50.2 & 73.6 & 51.9 & 74.5 & \tablepos{+1.7} & \tablepos{+0.9} \\
DISK        & 57.1 & 76.0 & 58.1 & 75.8 & \tablepos{+1.0} & \tableneg{-0.2} \\
ALIKED      & 61.6 & 78.0 & 61.9 & 76.9 & \tablepos{+0.3} & \tableneg{-1.1} \\
R2D2-SS     & 50.2 & 75.2 & 52.5 & 76.3 & \tablepos{+2.3} & \tablepos{+1.1} \\
R2D2-MS    & 41.4 & 69.4 & 44.6 & 70.4 & \tablepos{+3.2} & \tablepos{+1.0} \\
ORB-SS      & 35.5 & 64.0 & 38.1 & 64.8 & \tablepos{+2.6} & \tablepos{+0.8} \\
ORB-MS      & 23.1 & \textcolor{gray}{N/A} & 24.3 & \textcolor{gray}{N/A} & \tablepos{+1.2} & \textemdash \\
\bottomrule
\end{tabular}
\end{adjustbox}
\caption{\textbf{Image Matching Challenge 2021.} Stereo and multi-view (MV) mAA@10\degree on PhotoTourism. Shaded columns show fine-tuned models; $\Delta$ reports Ours minus Default. N/A denotes unavailable results.}
\label{table:IMC2021}
\end{table}

These results demonstrate that the proposed method generalizes across different detector and descriptor choices. In particular, the consistent improvements indicate that our approach is complementary to recent advances in local feature learning and can reliably enhance sparse matching performance in challenging cross-view scenarios such as air-to-ground localization.

\begin{table}
     \centering
     \small
     \renewcommand{\arraystretch}{1.12}
     \resizebox{\columnwidth}{!}{
     \begin{tabular}{@{}l ccc@{}}
     \toprule
      \textbf{Method} & \textbf{AUC@$5^{\circ}$}&\textbf{AUC@$10^{\circ}$}&\textbf{AUC@$20^{\circ}$}\\
        \midrule
        \multicolumn{4}{l}{\textit{Dense}} \\
        DKM~\tiny{CVPR'23}  & 65.0  & 77.2 & 85.7 \\
    RoMa~\tiny{CVPR'24} & 71.3 & 82.4 & 89.5 \\
    \midrule
    \multicolumn{4}{l}{\textit{Semi-Dense}} \\
    LoFTR~\cite{sun2021loftr}~\tiny{CVPR'21}& 21.5 & 33.8 & 45.6 \\
    ASpanFormer~\cite{chen2022aspanformer}~\tiny{ECCV'22} & 45.8 & 60.0 & 71.0 \\
    ELoFTR\tiny{CVPR'24} & 49.4 & 62.8 & 73.2 \\
    XFeat*\tiny{CVPR'24} & 12.0 & 19.2 & 27.9 \\
    RDD*\tiny{CVPR'25} & 43.8 & 55.3 & 64.9 \\
    \midrule
    \multicolumn{4}{l}{\textit{Sparse+NN}} \\
    ALIKED&12.0&17.8&25.8\\
    DeDoDe-V2-G& 31.5&45.3&58.3\\
    \midrule
  
    \multicolumn{4}{l}{\textit{Sparse + LightGlue}} \\
    SP+LG~\tiny{ICCV'23} & 47.9 & 62.7 & 73.9\\  
    RDD+LG~\tiny{ICCV'23} &  55.1 & 68.9 & 78.9  \\   
    \midrule
    \multicolumn{4}{l}{\textit{Sparse + ALIKED descriptor + LG}}\\
    DeDoDe-V2& 37.7&50.9&61.6\\
    DeDoDe-V2 (Ours)&45.0\,{\scriptsize\tablepos{+7.3}}&59.3\,{\scriptsize\tablepos{+8.4}}&70.8\,{\scriptsize\tablepos{+9.2}}\\
    RDD & 52.8&66.7&77.1\\
    RDD (Ours) & 54.0\,{\scriptsize\tablepos{+1.2}}&67.9\,{\scriptsize\tablepos{+1.2}}&77.8\,{\scriptsize\tablepos{+0.7}}\\
    DaD&53.8&68.0&78.5 \\
    DaD (Ours)&56.9\,{\scriptsize\tablepos{+3.1}}&70.9\,{\scriptsize\tablepos{+2.9}}&80.8\,{\scriptsize\tablepos{+2.3}} \\
    RaCo&54.4&69.1&79.2\\
    RaCo (Ours)&56.3\,{\scriptsize\tablepos{+1.9}}&70.6\,{\scriptsize\tablepos{+1.5}}&80.7\,{\scriptsize\tablepos{+1.5}}\\
    \midrule
    \multicolumn{4}{l}{\textit{Sparse + DeDoDe-G descriptor + LG}}\\
    
    DaD&56.8&70.8&80.7\\
    DaD (Ours)&\tablemetric{tablesecond}{58.4}\,{\scriptsize\tablepos{+1.6}}&\tablemetric{tablesecond}{72.6}\,{\scriptsize\tablepos{+1.8}}&\tablemetric{tablesecond}{82.2}\,{\scriptsize\tablepos{+1.5}}\\
    RaCo&56.3&71.2&81.6\\
    RaCo (Ours)&\tablemetric{tablebest}{58.5}\,{\scriptsize\tablepos{+2.2}}&\tablemetric{tablebest}{72.7}\,{\scriptsize\tablepos{+1.5}}&\tablemetric{tablebest}{82.7}\,{\scriptsize\tablepos{+1.1}}\\


     \bottomrule
     \end{tabular}
     }
     \caption{\textbf{Air-to-Ground matching.} Results on the benchmark~\cite{chen2025rdd} for the pose estimation task. Measured in AUC (higher is better). Green/bold and underlined values mark the best and second-best sparse matching results, respectively. Signed gains are shown beside fine-tuned results.}
     \label{table:air-ground2}
     \vspace{-0.05in}
 \end{table}
 
\subsection{Application to visual localization}
We evaluate our method on the widely-used Aachen Day-Night dataset~\cite{sattler2018benchmarking} and InLoc dataset~\cite{taira2018inloc}. 
The Aachen Day-Night dataset is a large-scale outdoor dataset for evaluating the visual localization methods across different times of the day. InLoc dataset is a challenging indoor visual localization dataset with high scene similarity and few texture.

 The accuracy is measured as the percentage of queries localized under three translation and rotation error thresholds (0.25 m/2$^\circ$, 0.5 m/5$^\circ$, 5m/10$^\circ$).
 The ground truth is hidden and the results are evaluated by the online server.

We use the hloc~\cite{sarlin2019coarse} toolbox for all experiments. For the Aachen dataset, a map with 3D point cloud and local features is constructed by COLMAP~\cite{schonberger2016structure} during the offline mapping stage. 
During the localization stage, 
the top 50 most similar database images to the query image are obtained with NetVLAD~\cite{arandjelovic2016netvlad}. Then 2D-2D matches are established by local matching. The final pose is solved with 2D-3D matches. For all detectors, we extract at most 4096 keypoints. For the InLoc dataset, there is no need for the SfM stage. 
The complete results are shown in \Cref{table:aachen_all_width}. We highlight the entries for which the accuracy difference exceeds 3\% between off-the-shelf model and our fine-tuned model.

\begin{table}[t]
\centering
\footnotesize
\setlength{\tabcolsep}{2.5pt}
\renewcommand{\arraystretch}{1.08}
\begin{adjustbox}{max width=\linewidth}
\begin{tabular}{@{}l c c c c c@{}}
\toprule
\multirow{2}{*}{Detector} & \multirow{2}{*}{Model} & \multicolumn{2}{c}{Aachen} & \multicolumn{2}{c}{InLoc} \\
\cmidrule(lr){3-4}\cmidrule(l){5-6}
& & Day & Night & DUC1 & DUC2 \\
\midrule
\multicolumn{6}{l}{\textit{DISK + LightGlue}} \\
SuperPoint & Base & 85.8/92.6 & 65.3/76.5 & 36.4/58.1 & 34.4/52.7 \\
& \cellcolor{tableheader}\textbf{Ours} & \cellcolor{tableheader}88.8/95.6 & \cellcolor{tableheader}84.7/92.9 & \cellcolor{tableheader}46.0/67.2 & \cellcolor{tableheader}47.3/67.9 \\
& $\Delta$ & {\scriptsize \tablepos{+3.0}/\tablepos{+3.0}} & {\scriptsize \tablepos{\textbf{+19.4}}/\tablepos{\textbf{+16.4}}} & {\scriptsize \tablepos{\textbf{+9.6}}/\tablepos{\textbf{+9.1}}} & {\scriptsize \tablepos{\textbf{+12.9}}/\tablepos{\textbf{+15.2}}} \\
\addlinespace[2pt]
ALIKED & Base & 87.5/95.0 & 77.6/86.7 & 37.4/58.1 & 42.0/58.8 \\
& \cellcolor{tableheader}\textbf{Ours} & \cellcolor{tableheader}88.2/95.4 & \cellcolor{tableheader}83.7/90.8 & \cellcolor{tableheader}47.5/65.7 & \cellcolor{tableheader}38.2/56.5 \\
& $\Delta$ & {\scriptsize \tablepos{+0.7}/\tablepos{+0.4}} & {\scriptsize \tablepos{\textbf{+6.1}}/\tablepos{\textbf{+4.1}}} & {\scriptsize \tablepos{\textbf{+10.1}}/\tablepos{\textbf{+7.6}}} & {\scriptsize \tableneg{\textbf{-3.8}}/\tableneg{-2.3}} \\
\addlinespace[2pt]
R2D2 & Base & 80.6/87.6 & 45.9/53.1 & 26.3/40.9 & 27.5/36.6 \\
& \cellcolor{tableheader}\textbf{Ours} & \cellcolor{tableheader}88.0/93.7 & \cellcolor{tableheader}78.6/88.8 & \cellcolor{tableheader}45.5/65.7 & \cellcolor{tableheader}36.6/54.2 \\
& $\Delta$ & {\scriptsize \tablepos{\textbf{+7.4}}/\tablepos{\textbf{+6.1}}} & {\scriptsize \tablepos{\textbf{+32.7}}/\tablepos{\textbf{+35.7}}} & {\scriptsize \tablepos{\textbf{+19.2}}/\tablepos{\textbf{+24.8}}} & {\scriptsize \tablepos{\textbf{+9.1}}/\tablepos{\textbf{+17.6}}} \\
\addlinespace[2pt]
ORB & Base & 76.8/86.2 & 41.8/52.0 & 28.8/44.9 & 15.3/32.8 \\
& \cellcolor{tableheader}\textbf{Ours} & \cellcolor{tableheader}82.0/91.3 & \cellcolor{tableheader}62.2/82.7 & \cellcolor{tableheader}37.4/54.0 & \cellcolor{tableheader}29.8/49.6 \\
& $\Delta$ & {\scriptsize \tablepos{\textbf{+5.2}}/\tablepos{\textbf{+5.1}}} & {\scriptsize \tablepos{\textbf{+20.4}}/\tablepos{\textbf{+30.7}}} & {\scriptsize \tablepos{\textbf{+8.6}}/\tablepos{\textbf{+9.1}}} & {\scriptsize \tablepos{\textbf{+14.5}}/\tablepos{\textbf{+16.8}}} \\
\midrule
\multicolumn{6}{l}{\textit{SuperPoint + LightGlue}} \\
DISK & Base & 85.6/93.6 & 81.6/90.8 & 39.4/55.6 & 38.9/61.8 \\
& \cellcolor{tableheader}\textbf{Ours} & \cellcolor{tableheader}85.7/94.5 & \cellcolor{tableheader}82.7/89.8 & \cellcolor{tableheader}42.4/62.6 & \cellcolor{tableheader}35.9/60.3 \\
& $\Delta$ & {\scriptsize \tablepos{+0.1}/\tablepos{+0.9}} & {\scriptsize \tablepos{+1.1}/\tableneg{-1.0}} & {\scriptsize \tablepos{+3.0}/\tablepos{\textbf{+7.0}}} & {\scriptsize \tableneg{-3.0}/\tableneg{-1.5}} \\
\addlinespace[2pt]
ALIKED & Base & 88.5/95.1 & 86.7/91.8 & 47.0/61.1 & 44.3/64.9 \\
& \cellcolor{tableheader}\textbf{Ours} & \cellcolor{tableheader}89.2/95.4 & \cellcolor{tableheader}84.7/90.8 & \cellcolor{tableheader}44.9/61.6 & \cellcolor{tableheader}45.8/69.5 \\
& $\Delta$ & {\scriptsize \tablepos{+0.7}/\tablepos{+0.3}} & {\scriptsize \tableneg{-2.0}/\tableneg{-1.0}} & {\scriptsize \tableneg{-2.1}/\tablepos{+0.5}} & {\scriptsize \tablepos{+1.5}/\tablepos{\textbf{+4.6}}} \\
\addlinespace[2pt]
R2D2 & Base & 88.6/94.9 & 87.8/91.8 & 45.5/66.7 & 49.6/70.2 \\
& \cellcolor{tableheader}\textbf{Ours} & \cellcolor{tableheader}88.7/94.8 & \cellcolor{tableheader}88.8/91.8 & \cellcolor{tableheader}50.0/69.7 & \cellcolor{tableheader}51.9/74.8 \\
& $\Delta$ & {\scriptsize \tablepos{+0.1}/\tableneg{-0.1}} & {\scriptsize \tablepos{+1.0}/\tablepos{0.0}} & {\scriptsize \tablepos{\textbf{+4.5}}/\tablepos{+3.0}} & {\scriptsize \tablepos{+2.3}/\tablepos{\textbf{+4.6}}} \\
\addlinespace[2pt]
ORB & Base & 79.9/90.4 & 67.3/83.7 & 40.4/57.6 & 35.1/54.2 \\
& \cellcolor{tableheader}\textbf{Ours} & \cellcolor{tableheader}80.8/90.7 & \cellcolor{tableheader}70.4/88.8 & \cellcolor{tableheader}39.4/58.1 & \cellcolor{tableheader}40.5/60.3 \\
& $\Delta$ & {\scriptsize \tablepos{+0.9}/\tablepos{+0.3}} & {\scriptsize \tablepos{\textbf{+3.1}}/\tablepos{\textbf{+5.1}}} & {\scriptsize \tableneg{-1.0}/\tablepos{+0.5}} & {\scriptsize \tablepos{\textbf{+5.4}}/\tablepos{\textbf{+6.1}}} \\
\midrule
\multicolumn{6}{l}{\textit{ALIKED + LightGlue}} \\
SuperPoint & Base & 89.2/95.5 & 84.7/92.9 & 45.5/66.7 & 45.8/69.5 \\
& \cellcolor{tableheader}\textbf{Ours} & \cellcolor{tableheader}88.6/95.9 & \cellcolor{tableheader}86.7/94.9 & \cellcolor{tableheader}44.9/65.7 & \cellcolor{tableheader}42.7/74.8 \\
& $\Delta$ & {\scriptsize \tableneg{-0.6}/\tablepos{+0.4}} & {\scriptsize \tablepos{+2.0}/\tablepos{+2.0}} & {\scriptsize \tableneg{-0.6}/\tableneg{-1.0}} & {\scriptsize \tableneg{\textbf{-3.1}}/\tablepos{\textbf{+5.3}}} \\
\addlinespace[2pt]
DISK & Base & 86.3/94.2 & 84.7/90.8 & 42.9/60.6 & 37.4/60.3 \\
& \cellcolor{tableheader}\textbf{Ours} & \cellcolor{tableheader}85.8/94.3 & \cellcolor{tableheader}82.7/89.8 & \cellcolor{tableheader}43.9/60.1 & \cellcolor{tableheader}36.6/65.6 \\
& $\Delta$ & {\scriptsize \tableneg{-0.5}/\tablepos{+0.1}} & {\scriptsize \tableneg{-2.0}/\tableneg{-1.0}} & {\scriptsize \tablepos{+1.0}/\tableneg{-0.5}} & {\scriptsize \tableneg{-0.8}/\tablepos{\textbf{+5.3}}} \\
\addlinespace[2pt]
R2D2 & Base & 88.2/94.5 & 84.7/91.8 & 51.0/71.2 & 49.6/73.3 \\
& \cellcolor{tableheader}\textbf{Ours} & \cellcolor{tableheader}89.8/94.9 & \cellcolor{tableheader}88.8/92.9 & \cellcolor{tableheader}50.0/71.2 & \cellcolor{tableheader}53.4/73.3 \\
& $\Delta$ & {\scriptsize \tablepos{+1.6}/\tablepos{+0.4}} & {\scriptsize \tablepos{\textbf{+4.1}}/\tablepos{+1.1}} & {\scriptsize \tableneg{-1.0}/\tablepos{0.0}} & {\scriptsize \tablepos{\textbf{+3.8}}/\tablepos{0.0}} \\
\addlinespace[2pt]
ORB & Base & 82.3/91.7 & 69.4/89.8 & 40.9/59.6 & 38.2/60.3 \\
& \cellcolor{tableheader}\textbf{Ours} & \cellcolor{tableheader}82.9/91.3 & \cellcolor{tableheader}71.4/88.8 & \cellcolor{tableheader}41.9/59.1 & \cellcolor{tableheader}39.7/64.9 \\
& $\Delta$ & {\scriptsize \tablepos{+0.6}/\tableneg{-0.4}} & {\scriptsize \tablepos{+2.0}/\tableneg{-1.0}} & {\scriptsize \tablepos{+1.0}/\tableneg{-0.5}} & {\scriptsize \tablepos{+1.5}/\tablepos{\textbf{+4.6}}} \\
\bottomrule
\end{tabular}
\end{adjustbox}
\caption{\textbf{Visual localization.} Recall on Aachen Day-Night and InLoc. Base: official model; Ours: fine-tuned. $\Delta$ reports signed changes in percentage points; changes exceeding 3 points in magnitude are bold.}
\label{table:aachen_all_width}
\vspace{-0.05in}
\end{table}

DISK descriptors perform poorly when coupled with other detectors when using the official LightGlue model. When using R2D2 or ORB as detector, it localizes fewer than $50\%$ queries under (0.25 m/2$^\circ$) for Aachen night queries. After fine-tuning the matching model with our method, it shows significant performance gain for all detectors. 
For R2D2 and ORB keypoints, the absolute gain is 32.7\% and 20.4\% under (0.25 m/2$^\circ$). 
We also obtain significant improvement when coupling DISK matcher with other detectors on the InLoc dataset, with the accuracy improvement under 0.25 m/2$^\circ$ ranging from 8.6\% to 19.2\%. The only exception occurs when using ALIKED keypoints with DISK matcher for duc2, with minor degradation of 3.8\%. 

SuperPoint descriptors coupled with all learned detectors successfully localizes $>$82\%  queries under (0.25 m/2$^\circ$) for Aachen night queries. Interestingly, SuperPoint descriptors were originally trained only for SuperPoint keypoints, yet they generalize well to other detectors. When using R2D2 features extracted at single-scale, it obtains the best results for Aachen and InLoc datasets. This is contrary to previous work~\cite{oblivious} that suggests the SuperGlue model trained for SuperPoint will not generalize to R2D2 keypoints. 
With ORB keypoints, we localize 70.4\% night queries under (0.25 m/2$^\circ$), which indicates the strength of attention-based method for matching ultra-fast binary feature detectors.

On the Aachen Day-Night benchmark, ALIKED descriptors perform on par with SuperPoint. Fine-tuning yields the largest gain for R2D2 keypoints (+4.1 at 0.25m / 2$^\circ$). ORB keypoints, when coupled with ALIKED descriptors, achieve record-high accuracy of 71.4\%.

\section{Conclusion}
\label{sec:conclusion}
We revisit the application of transformer-based image matching models on different local features. An effective approach is introduced to remove nearby keypoints, leading to a significant performance gain. To better understand the roles of detectors and descriptors, we conduct a decoupled analysis within the attention-based image matching framework. Furthermore, we propose a novel fine-tuning method to adapt existing specialized models into detector-agnostic models. Experimental results on public benchmarks across various tasks demonstrate the effectiveness of our approach. We believe our work provides valuable insights for developing improved sparse image matching models and local features.
\section{Appendix}

\subsection{Test-time ensemble}
\begin{figure}[b]
\centering
\includegraphics[width=\linewidth]{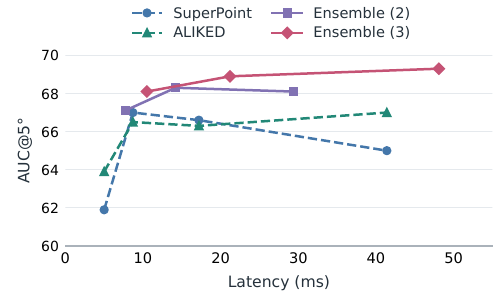}
\caption{\textbf{Accuracy--latency trade-off.} Ensemble (2) combines SuperPoint and ALIKED; Ensemble (3) additionally uses SIFT, with a single matcher. Individual detectors use 1K/2K/4K/8K keypoints; ensembles use 1K/2K/4K per detector.}
\label{fig:ensemble}
\end{figure}

Our methods enables test-time ensemble with multiple detectors \textbf{with single model}, which show consistent gain with more detectors included as shown in \cref{fig:ensemble}. 
Training with cross-modality data to make existing matching models work across various modalities is promising future work since we have shown that attention-based methods indeed suffer from detector bias, as shown in concurrent works\cite{he2025matchanything,xue2025matchatowardsmatching}.

\subsection{Additional Details}
Each model is fine-tuned for 10 epochs. The initial learning rate is set at 5e-5 for the first 5 epochs. The learning rate is exponentially reduced to 5e-6 after 10 epochs.
The fine-tuning process for each model takes about 1.5 days. 

\begin{figure}[t]
\centering
\includegraphics[width=\linewidth]{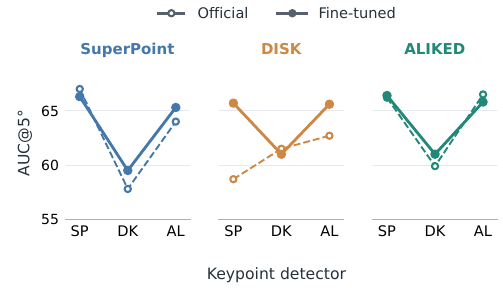}
\caption{\textbf{Transfer across keypoint detectors.} Each panel uses one descriptor--matcher pair. Dashed/open markers show official models; solid/filled markers show fine-tuned models. SP, DK, and AL denote SuperPoint, DISK, and ALIKED detectors. Results are AUC@5\degree on MegaDepth-1500.}
\label{tab:tuned}
\end{figure}

The relative pose estimation accuracy on the MegaDepth-1500 dataset using SuperPoint, DISK, and ALIKED after fine tuning are shown in \cref{tab:tuned}. With the official model, matching the same keypoints with different descriptors results in performances degradation. For example, matching SuperPoint keypoints with DISK descriptors obtains 58.7\% under AUC@5\degree, compared to 67.0\% with SuperPoint descriptors. After fine-tuning, the performance gap is largely closed, with the AUC@5\degree of 65.7\%. The AUC is comparable for the same detectors no matter what descriptors or matchers are used. 

For LightGlue models trained on R2D2, XFeat, and ORB, the original descriptors are projected to 256 dimensions to maintain the same network capacity. Early exit for LightGlue models are disabled through all experiments. 

The maximum image size for multi-scale extraction is the same as single-scale extraction, 1600 for Aachen Day-Night dataset. The minimum size is set to 256 following previous work~\cite{oblivious}.

For ORB features, we extract 5 scales by default. 

For Dedode in Kornia, version 0.7.4 was used, in which NMS is absent. We have submitted a fix to Kornia.\footnote{See the PR on GitHub: \href{https://github.com/kornia/kornia/pull/3153}{kornia/kornia\#3153}}

The experimental setup for the Air-to-Ground benchmark follows original  paper~\cite{chen2025rdd}. For keypoint extraction, we use a maximum size of 1600 and 4096 keypoints. Notably, for the DeDoDe-G descriptors, we resize the input image to ensure that its dimensions are divisible by 14, as specified by the method's requirements.

The latency reported in \cref{fig:ensemble} was measured on a machine equipped with a single GTX 4090 GPU.

\subsection{Visualization}
\label{sec:appendix_section}
\begin{figure*}[t]
\centering
\begingroup
\setlength{\tabcolsep}{3pt}
\renewcommand{\arraystretch}{1.0}
\begin{adjustbox}{max width=\textwidth}
\begin{tabular}{@{}c cc@{\hspace{10pt}}cc@{}}
& \multicolumn{2}{c}{\small\textbf{Example 1}} & \multicolumn{2}{c}{\small\textbf{Example 2}} \\
\cmidrule(lr){2-3}\cmidrule(l){4-5}
& \small Official & \small\textbf{Ours} & \small Official & \small\textbf{Ours} \\[4pt]
\raisebox{-.5\height}{\rotatebox{90}{\small\textbf{DISK + LG}}} & \raisebox{-.5\height}{\includegraphics[height=2.2in]{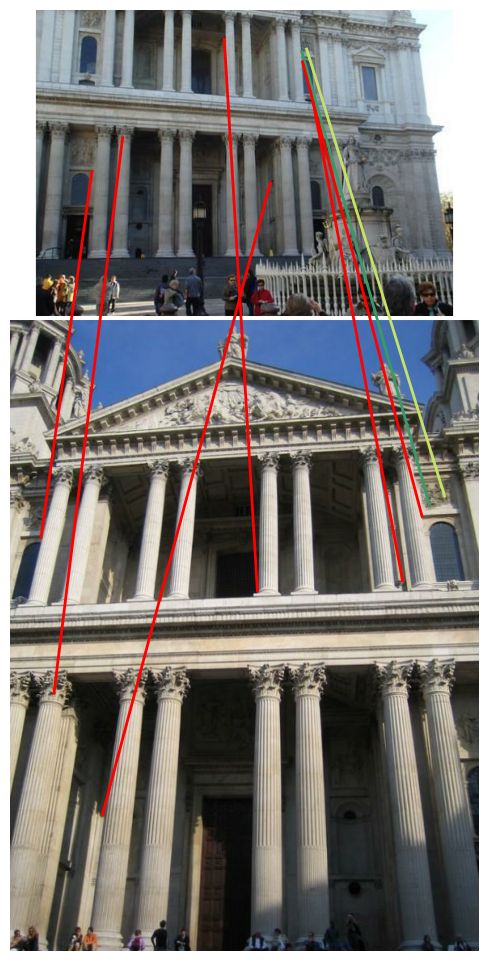}} & \raisebox{-.5\height}{\includegraphics[height=2.2in]{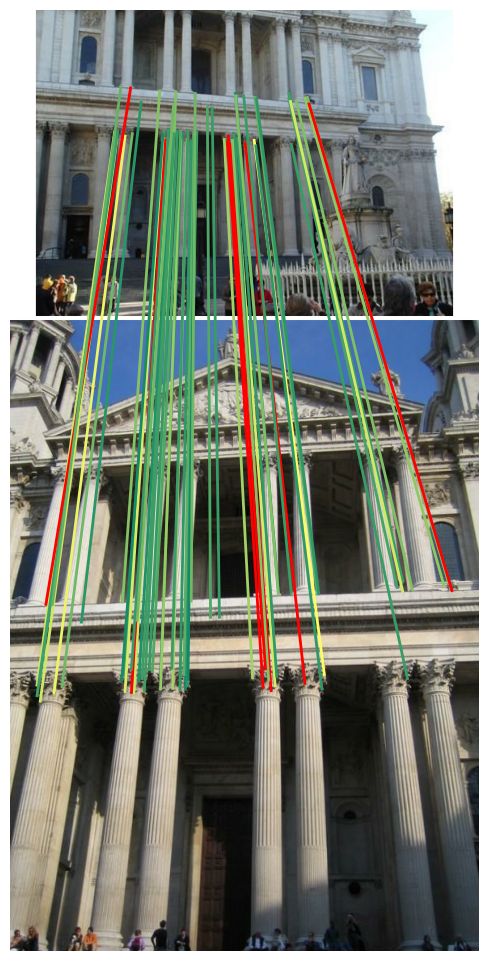}} & \raisebox{-.5\height}{\includegraphics[height=2.2in]{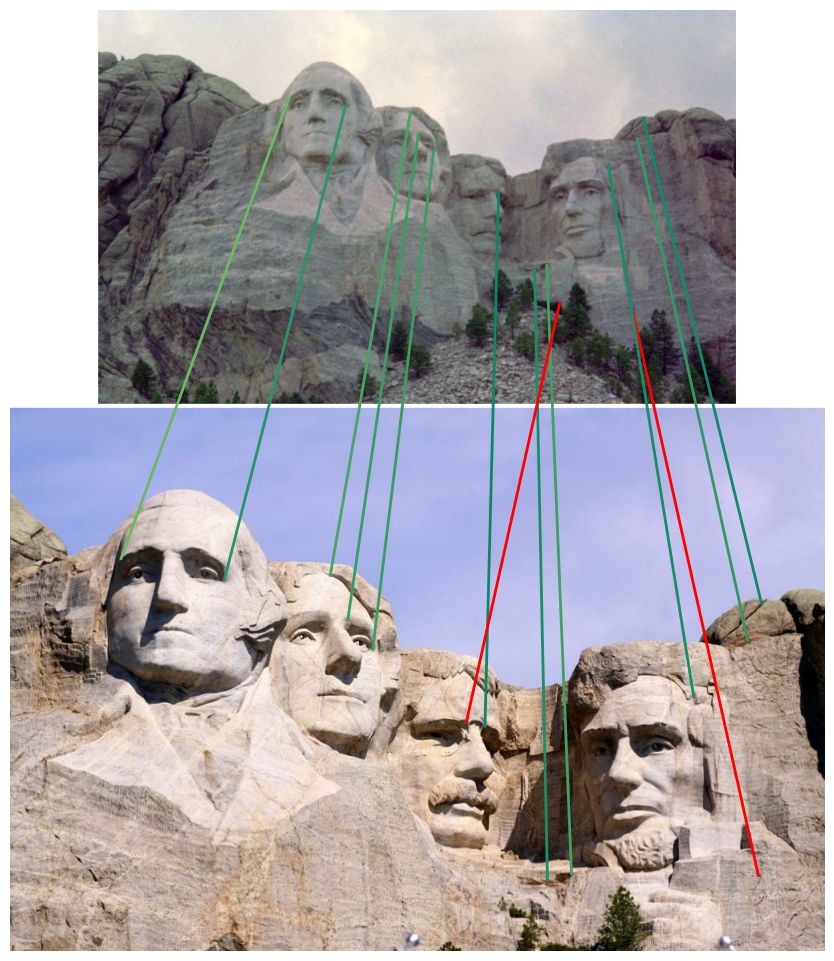}} & \raisebox{-.5\height}{\includegraphics[height=2.2in]{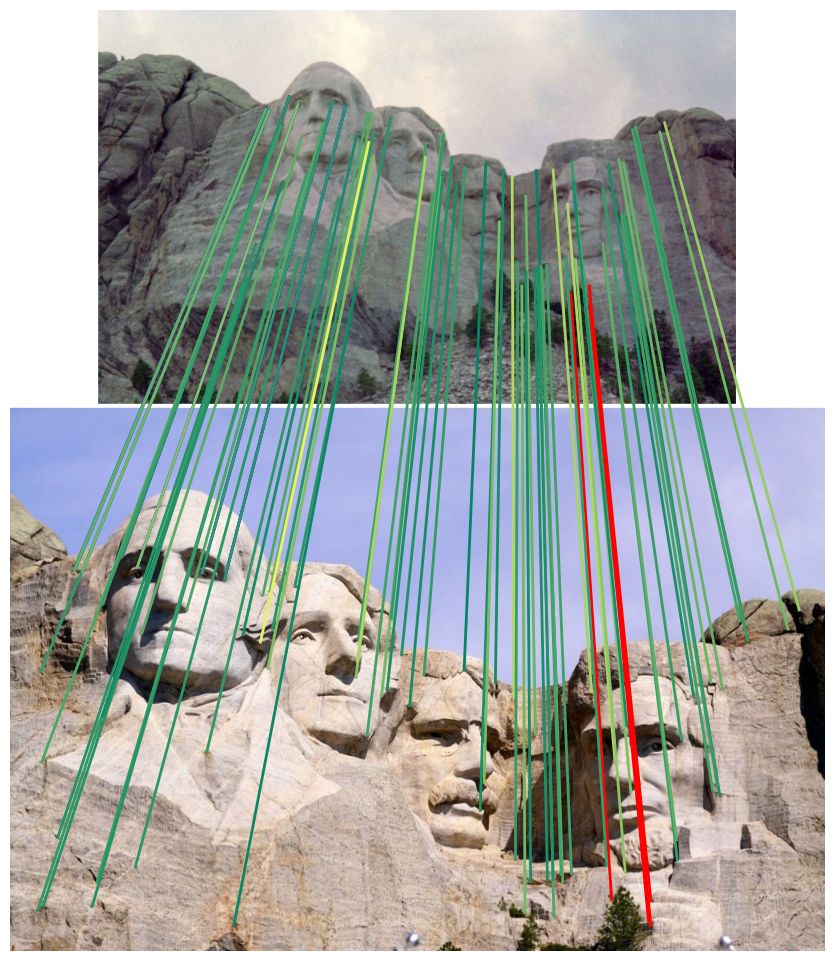}} \\[6pt]
\raisebox{-.5\height}{\rotatebox{90}{\small\textbf{SuperPoint + LG}}} & \raisebox{-.5\height}{\includegraphics[height=2.2in]{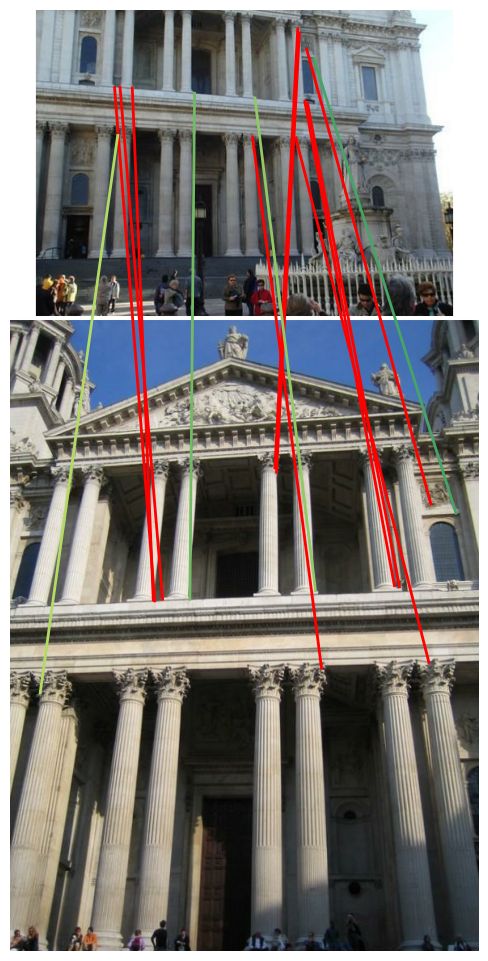}} & \raisebox{-.5\height}{\includegraphics[height=2.2in]{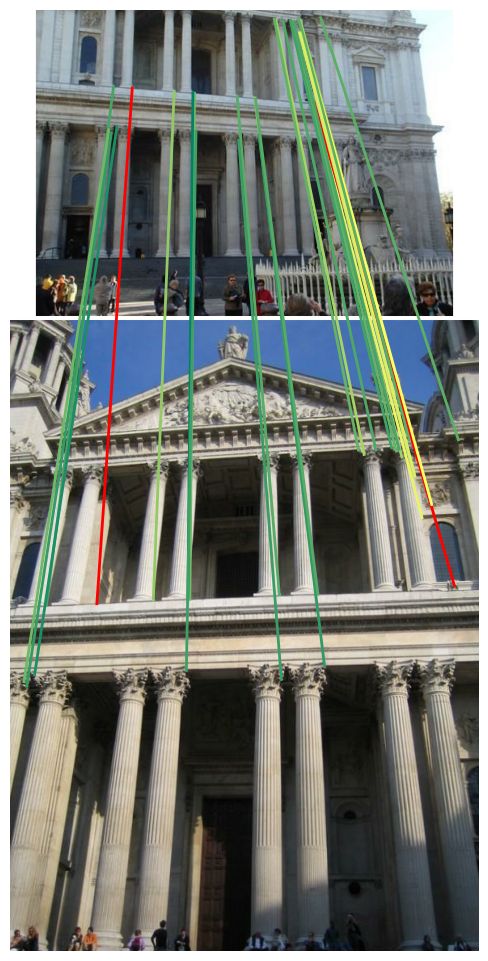}} & \raisebox{-.5\height}{\includegraphics[height=2.2in]{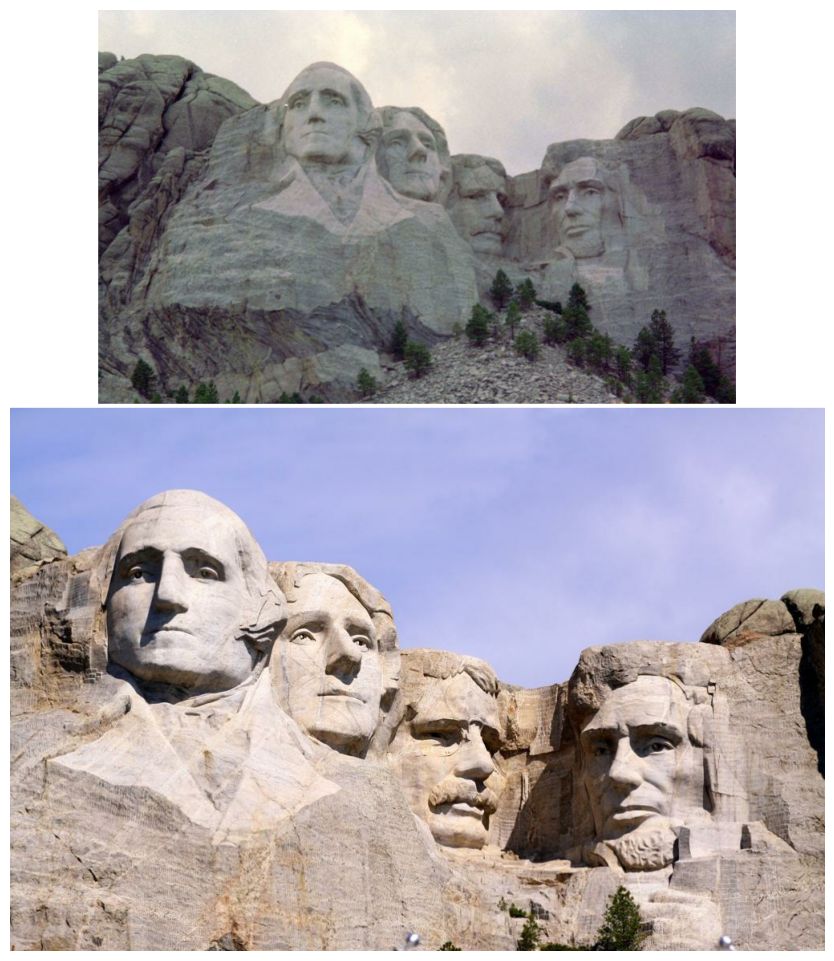}} & \raisebox{-.5\height}{\includegraphics[height=2.2in]{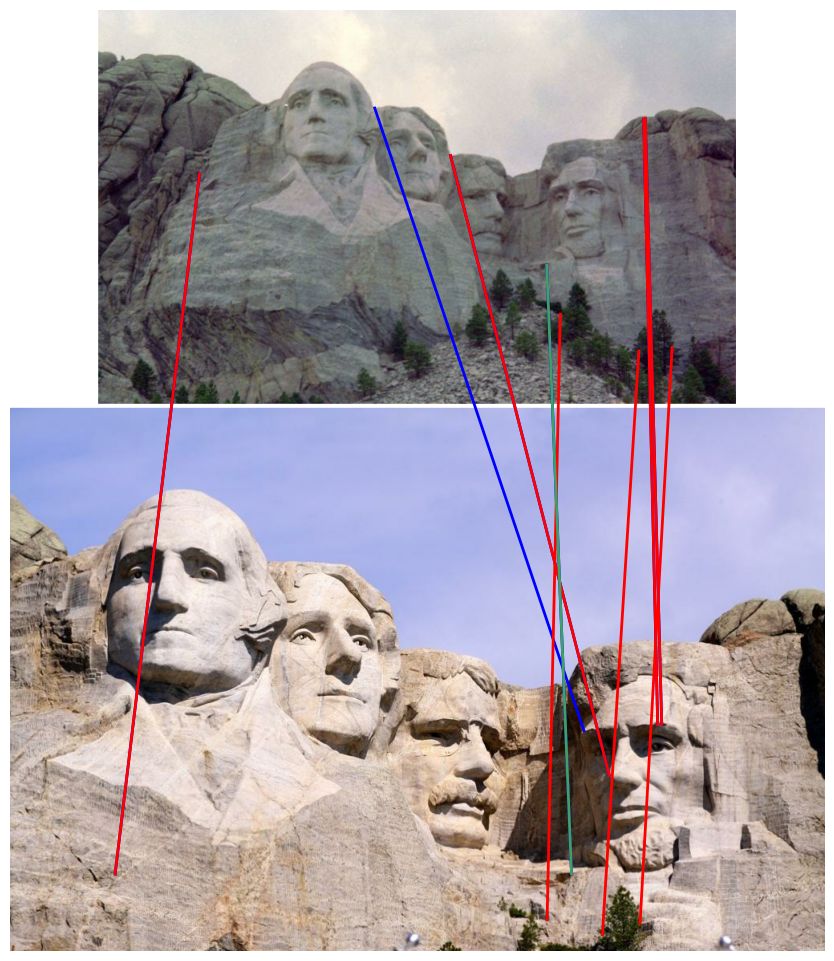}} \\[6pt]
\raisebox{-.5\height}{\rotatebox{90}{\small\textbf{ALIKED + LG}}} & \raisebox{-.5\height}{\includegraphics[height=2.2in]{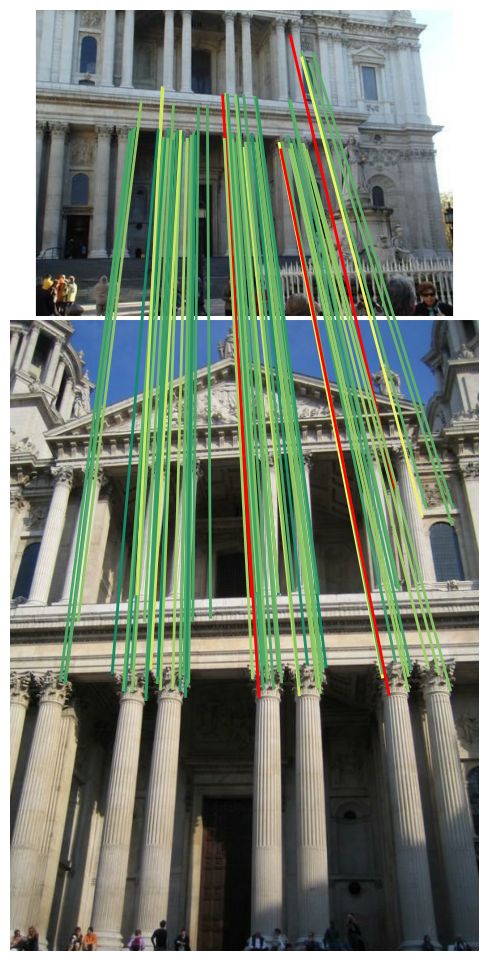}} & \raisebox{-.5\height}{\includegraphics[height=2.2in]{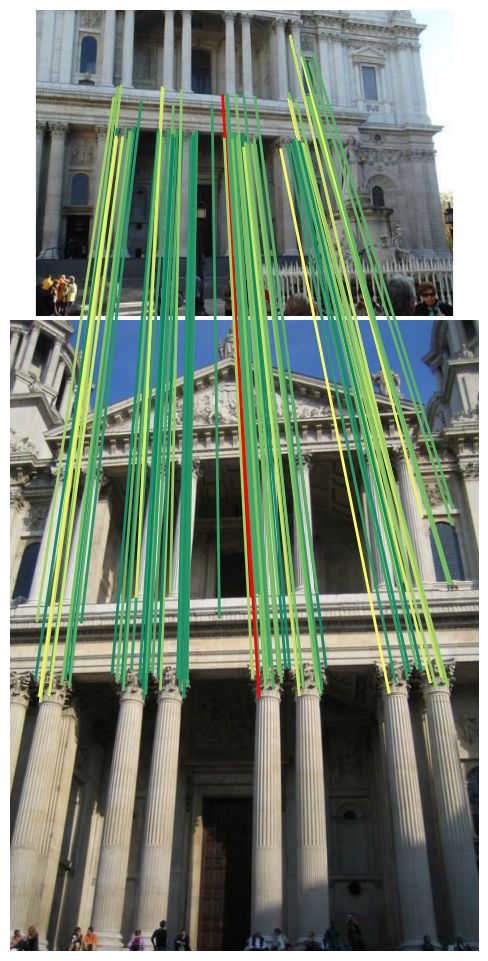}} & \raisebox{-.5\height}{\includegraphics[height=2.2in]{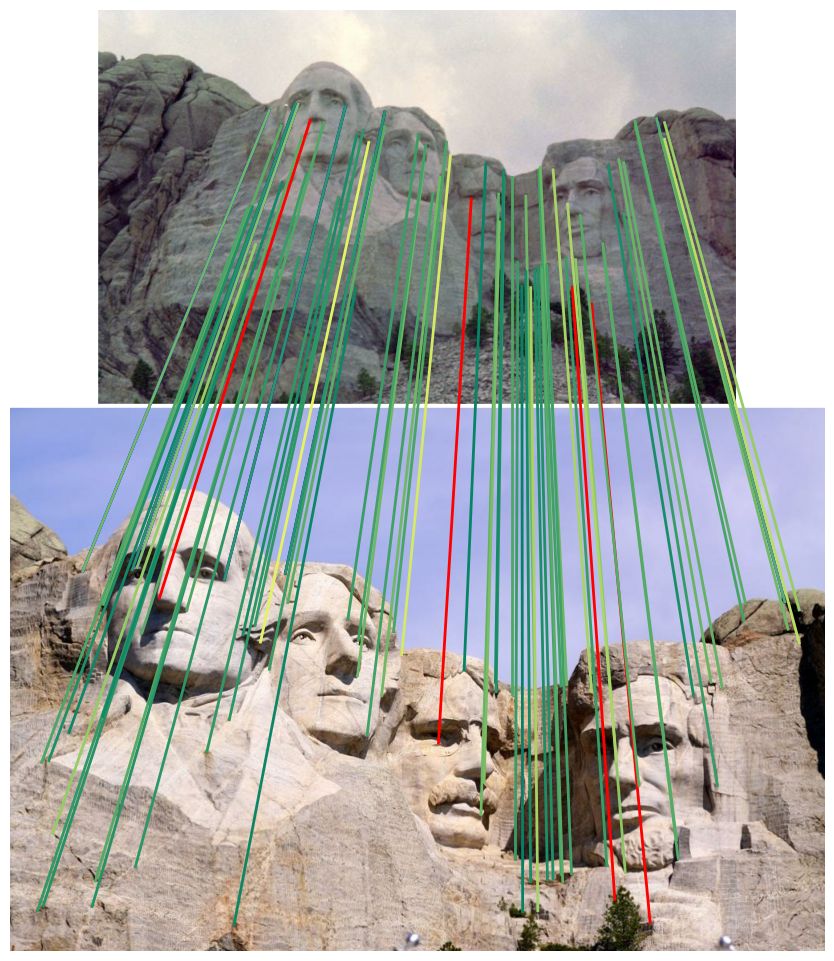}} & \raisebox{-.5\height}{\includegraphics[height=2.2in]{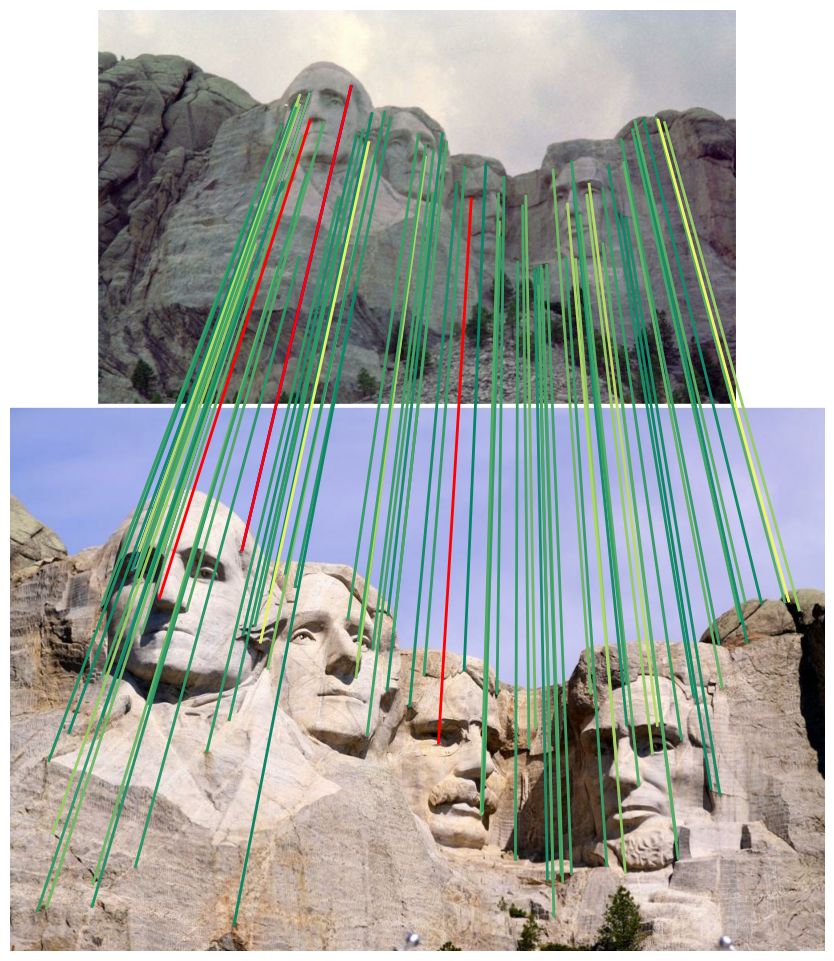}} \\
\end{tabular}
\end{adjustbox}
\endgroup
\caption{\textbf{Qualitative stereo matching on IMC2021.} ORB keypoints matched with DISK, SuperPoint, and ALIKED descriptors (rows). For each example, adjacent columns compare the official LightGlue model with our fine-tuned model.}
\label{fig:vis_imc_stereo}
\end{figure*}

We show image matching results for stereo task on IMC2021~\cite{imc2021} dataset in \cref{fig:vis_imc_stereo}. All keypoints are detected by ORB~\cite{rublee2011orb} extracted at single scale. They are matched with LightGlue models trained for DISK, SuperPoint and ALIKED descriptors. The results with off-the-shelf models are shown on the top, while the results with our models are shown below. 
Our models obtain significantly more correct matches compared with the off-the-shelf models~\cite{lindenberger2023lightglue}, suggesting the importance of our fine-tuning method. Compared to LightGlue model for DISK and SuperPoint, the model for ALIKED descriptors obtains most correct matches.

\begin{figure*}[t]
\centering
\subfloat{\includegraphics[width=0.24\linewidth,height=2.in]{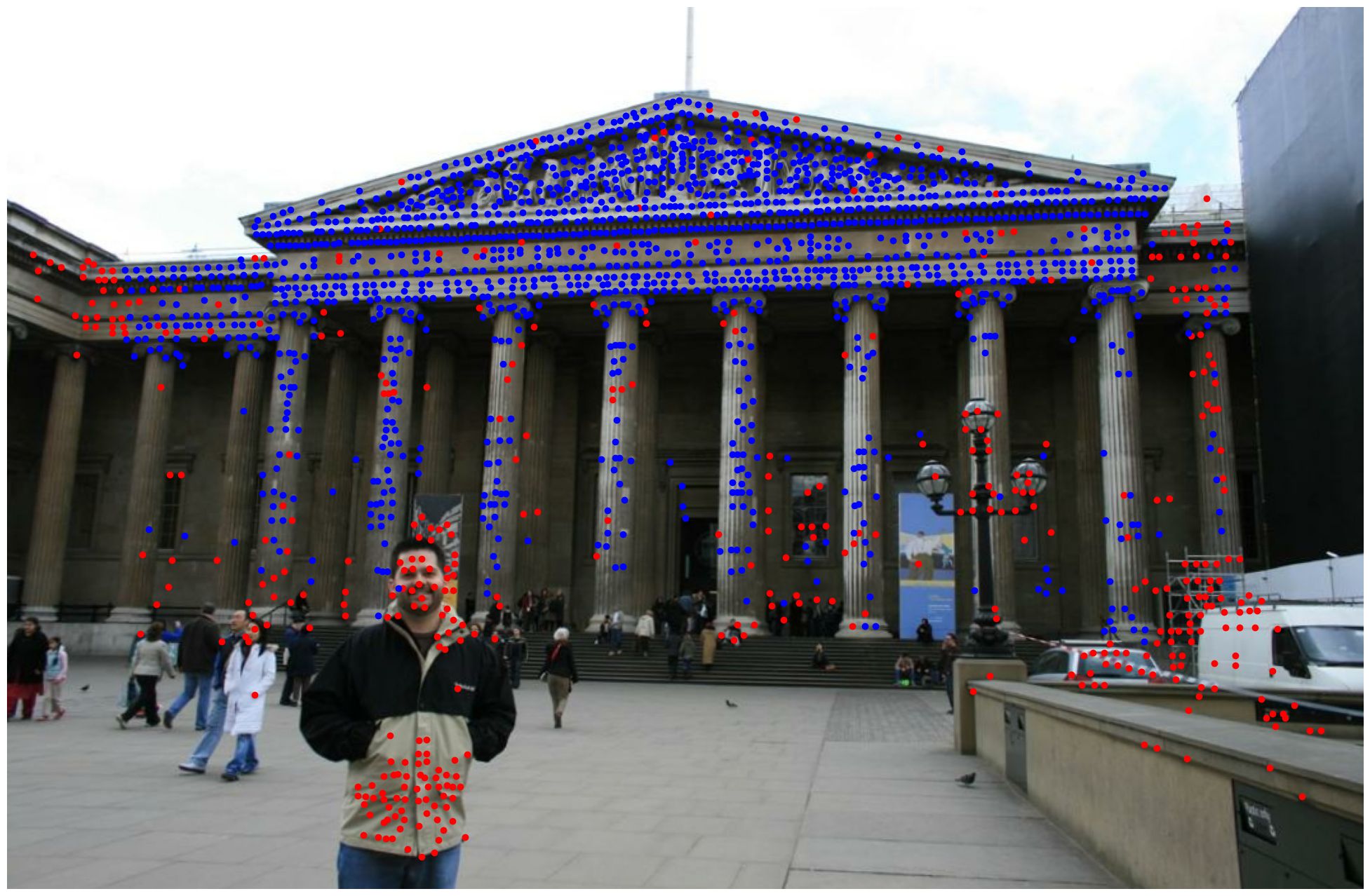}}
\subfloat{\includegraphics[width=0.24\linewidth,height=2.in]{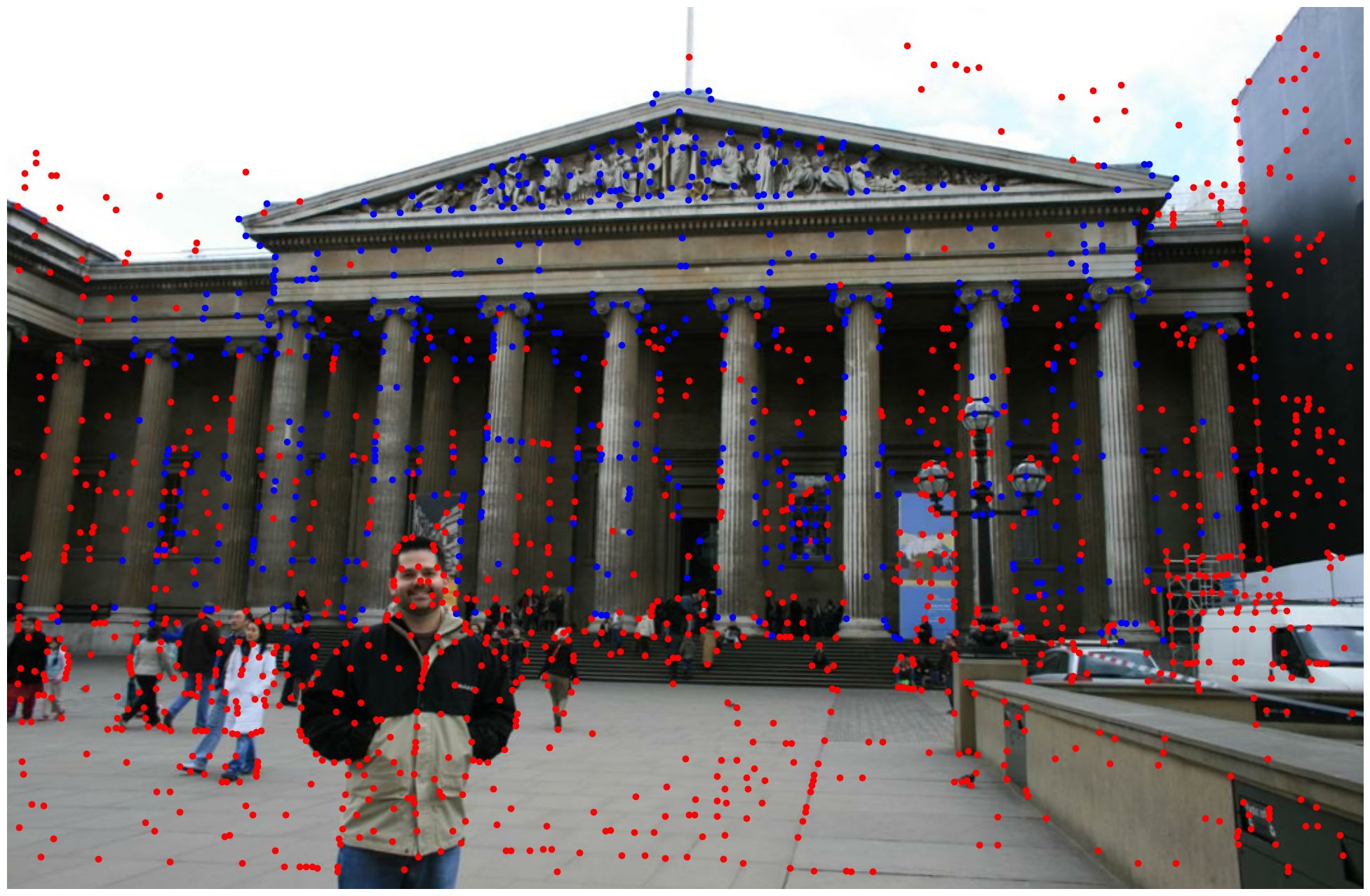}}
\subfloat{\includegraphics[width=0.24\linewidth,height=2.in]{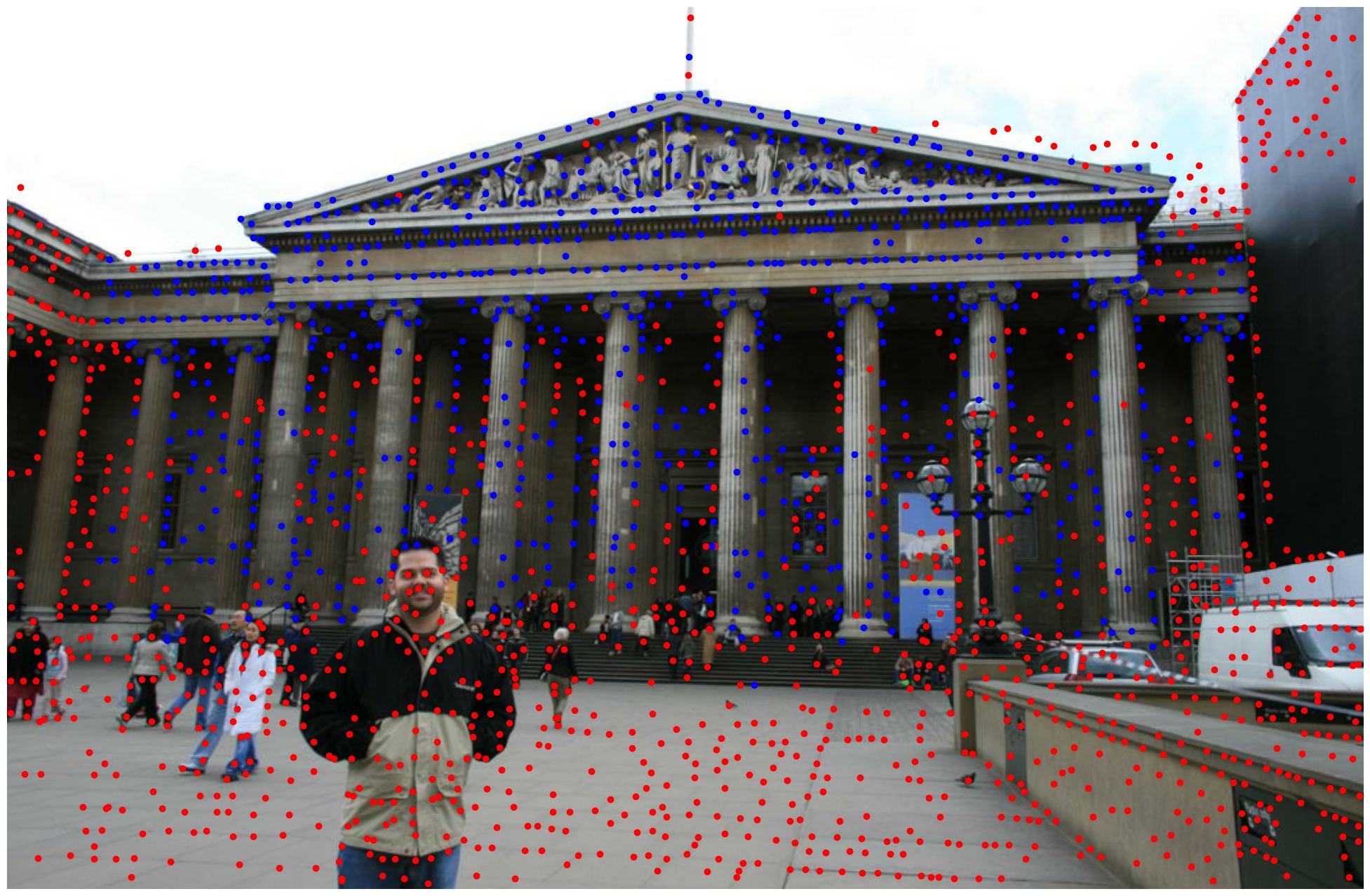}}
\subfloat{\includegraphics[width=0.24\linewidth,height=2.in]{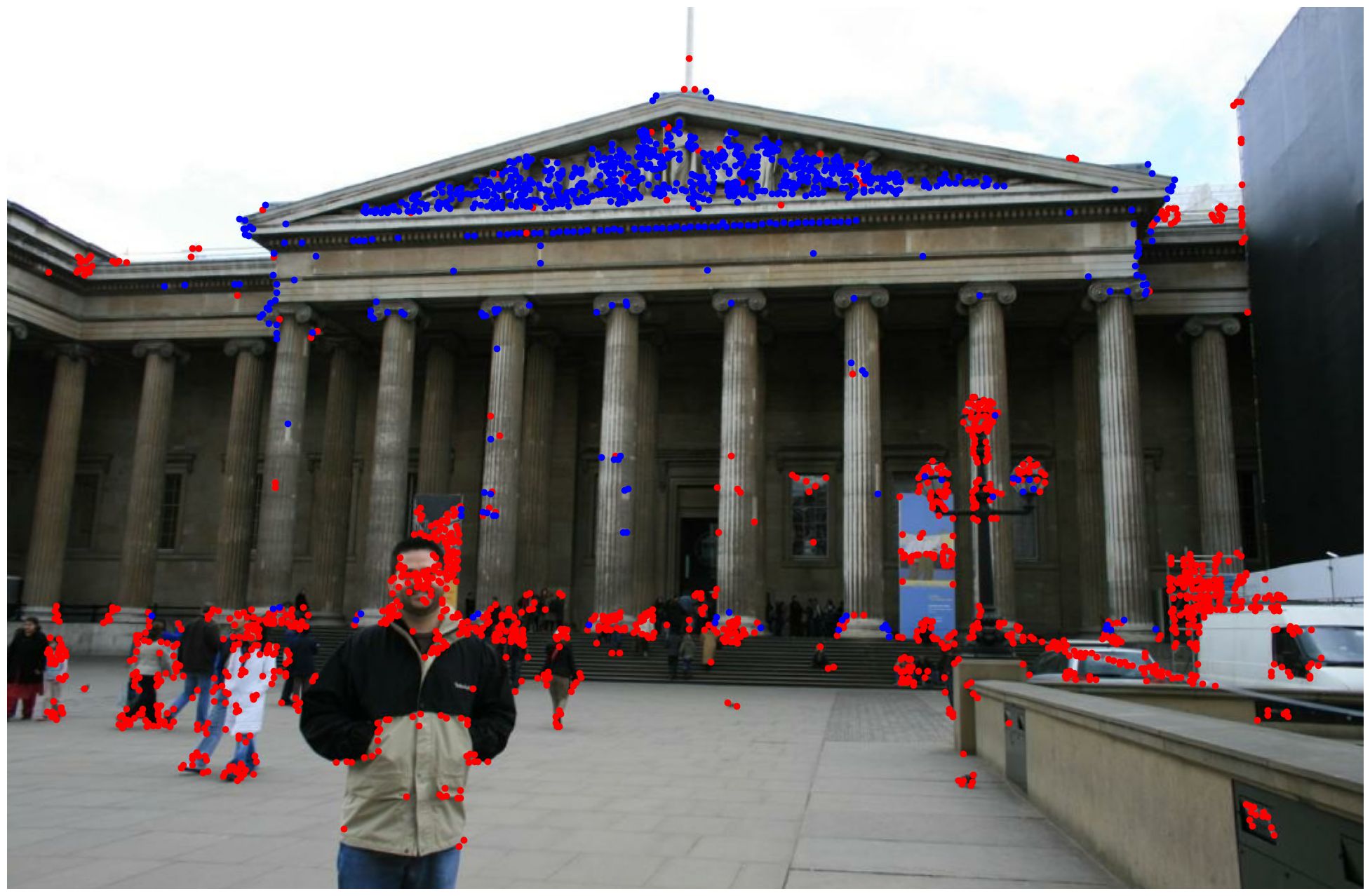}}

\subfloat{\includegraphics[width=0.24\linewidth,height=2.in]{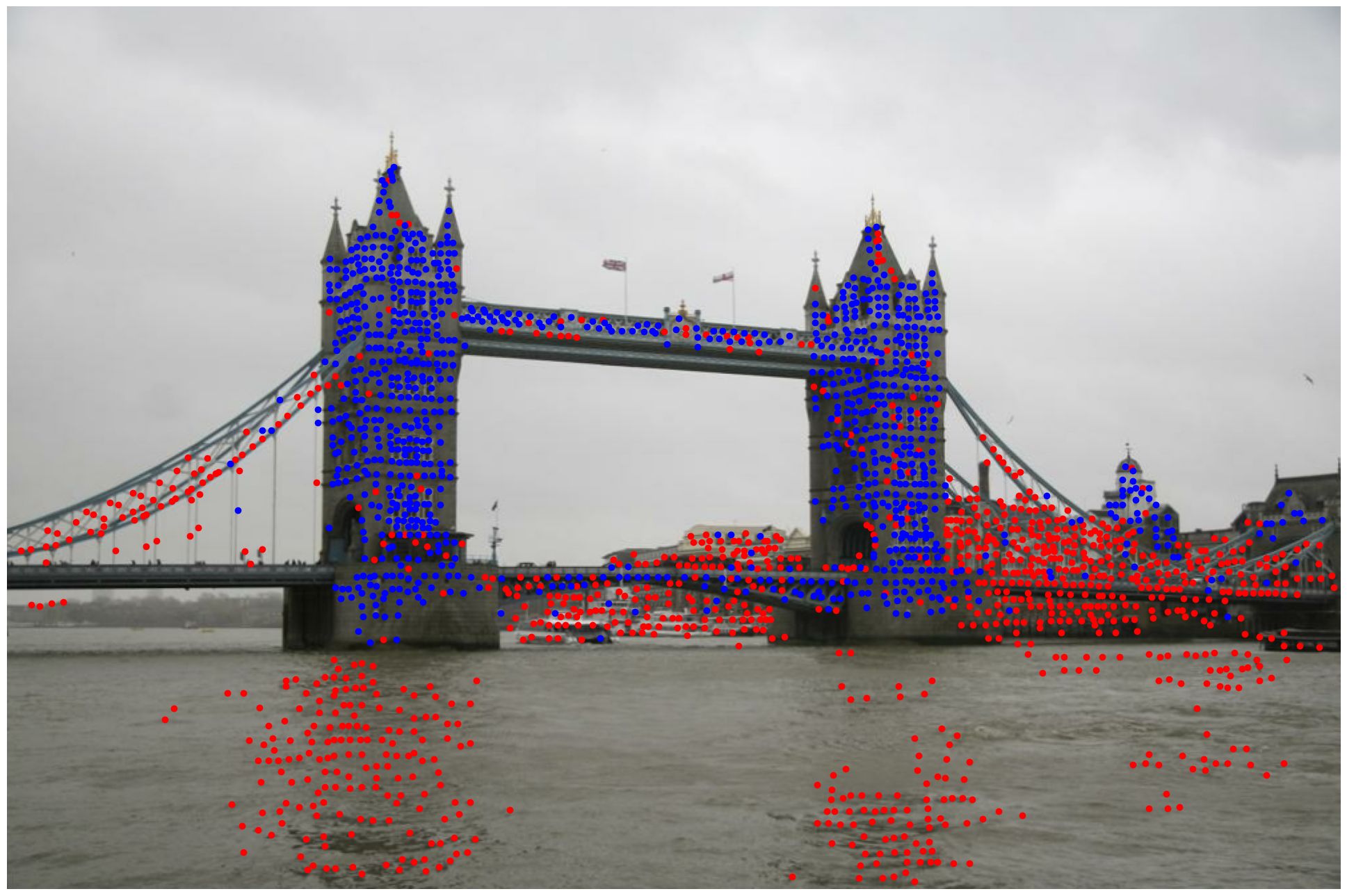}}
\subfloat{\includegraphics[width=0.24\linewidth,height=2.in]{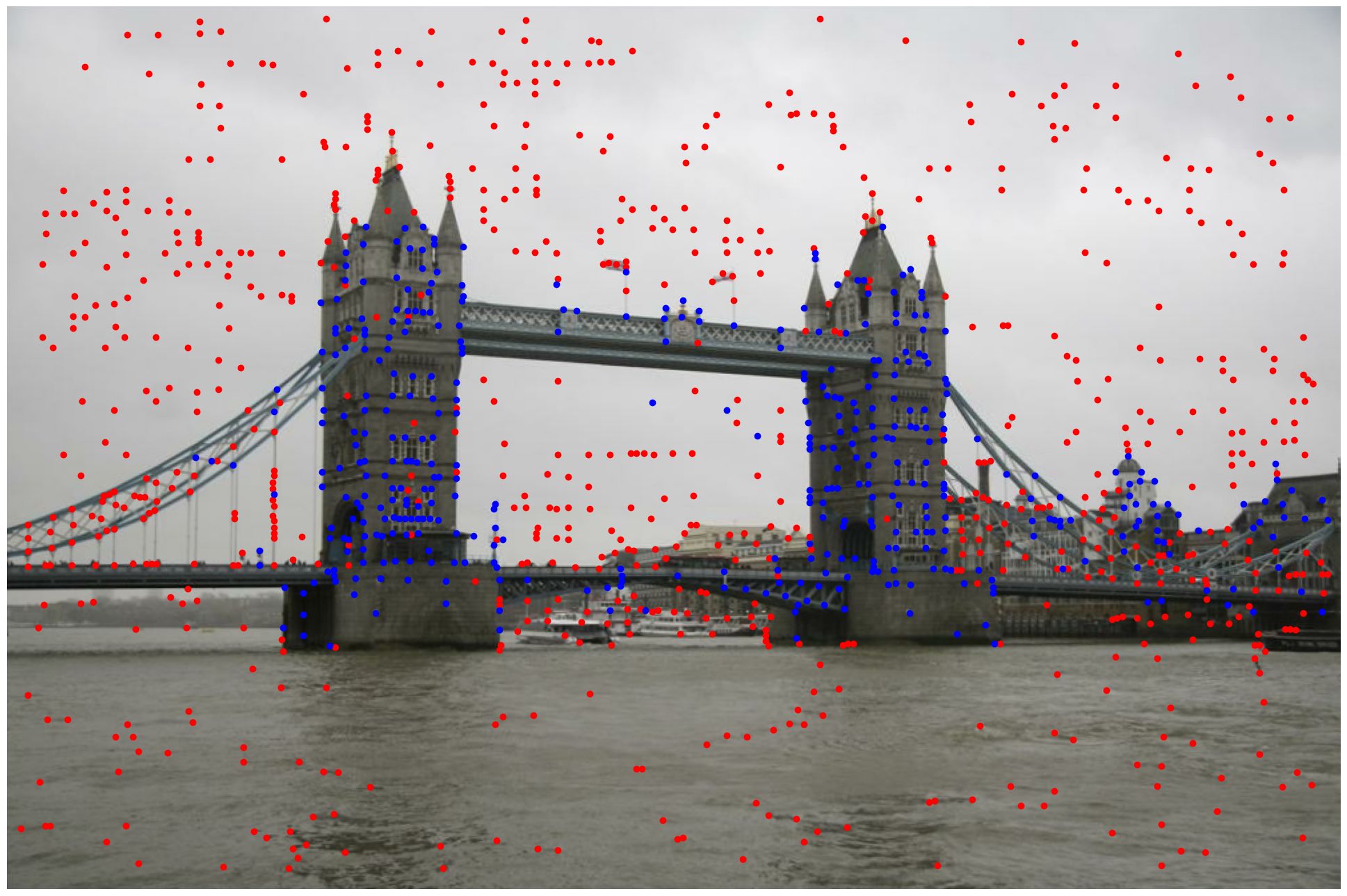}}
\subfloat{\includegraphics[width=0.24\linewidth,height=2.in]{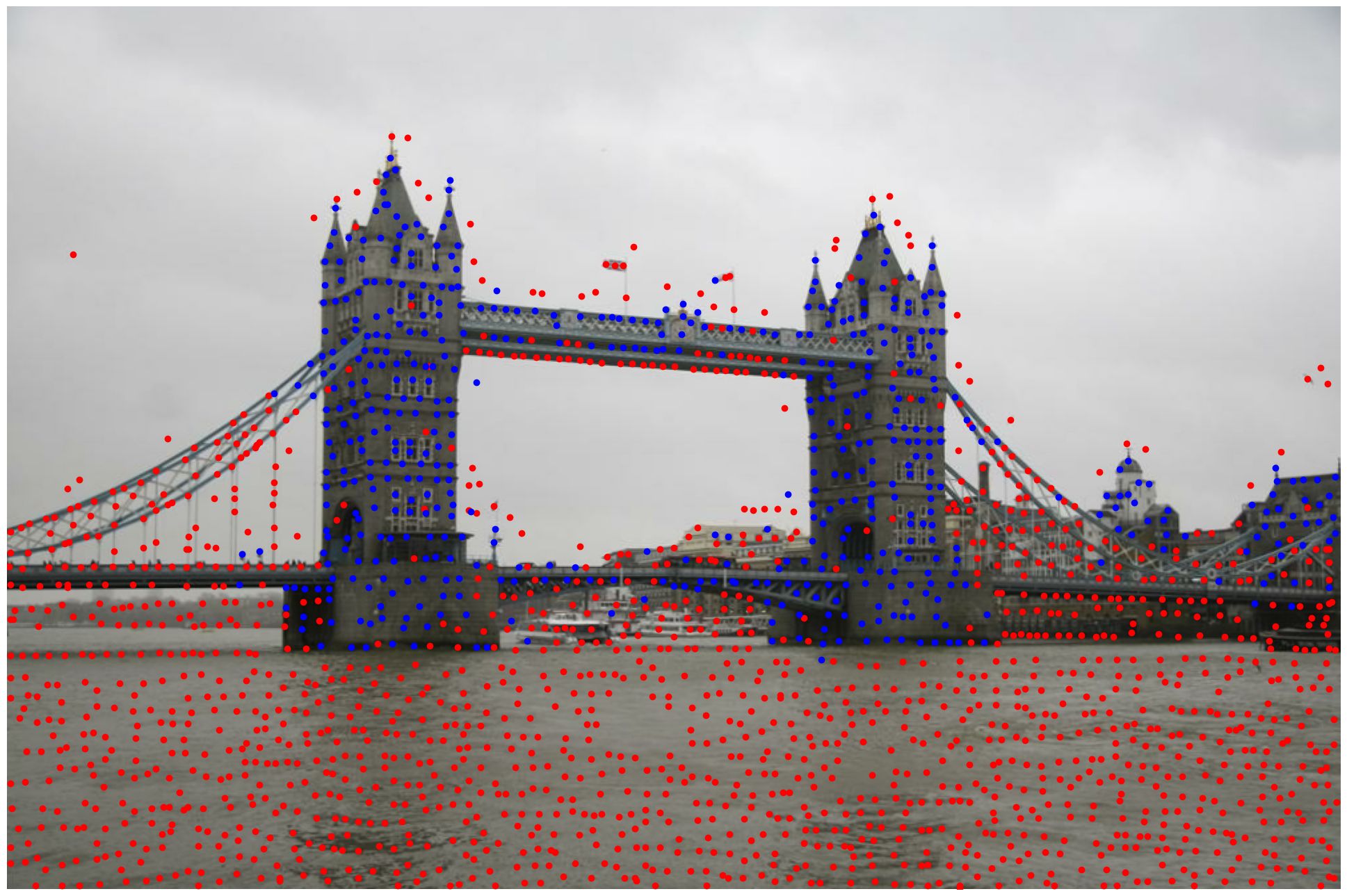}}
\subfloat{\includegraphics[width=0.24\linewidth,height=2.in]{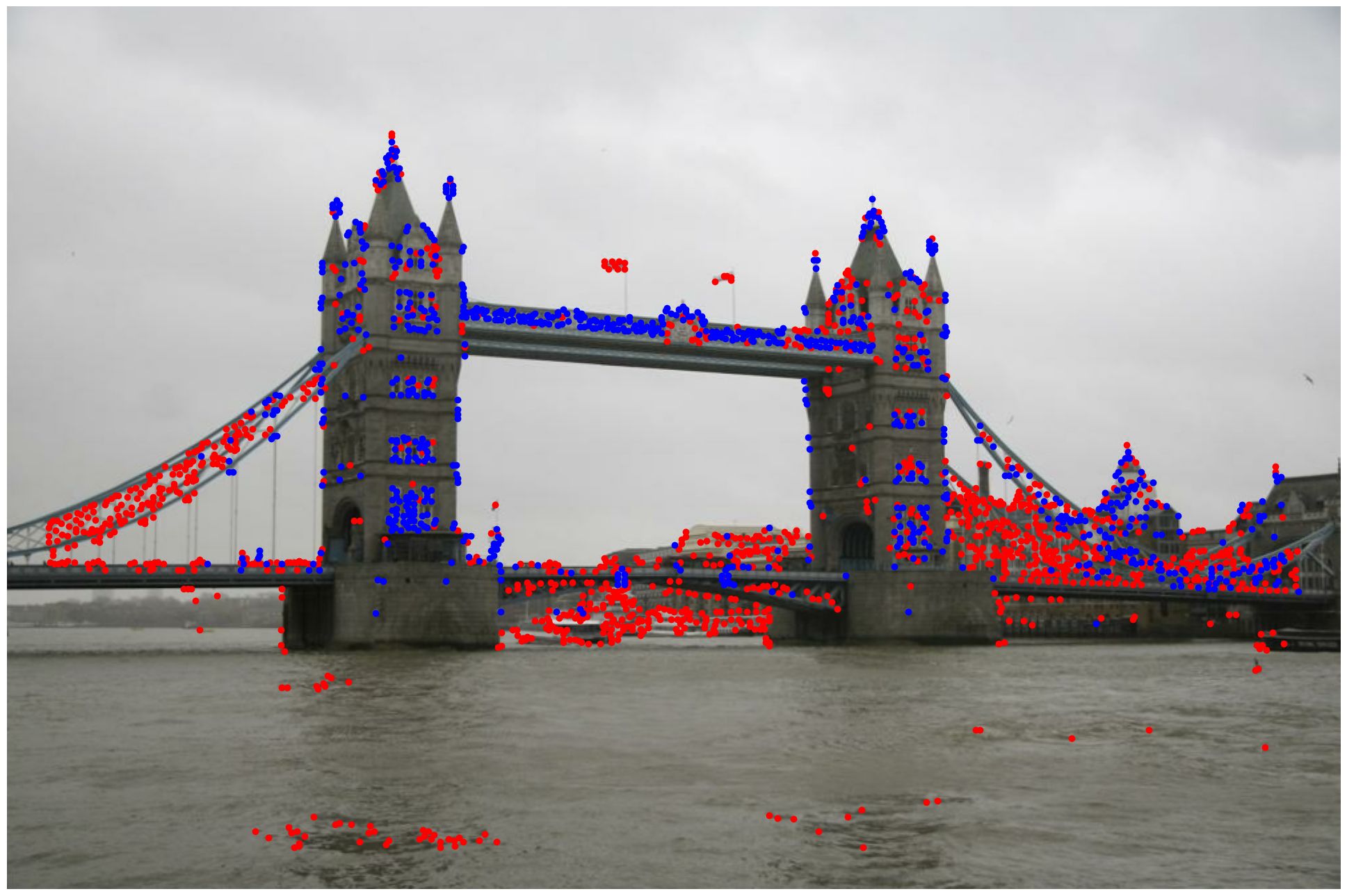}}

\subfloat{\includegraphics[width=0.24\linewidth,height=2.in]{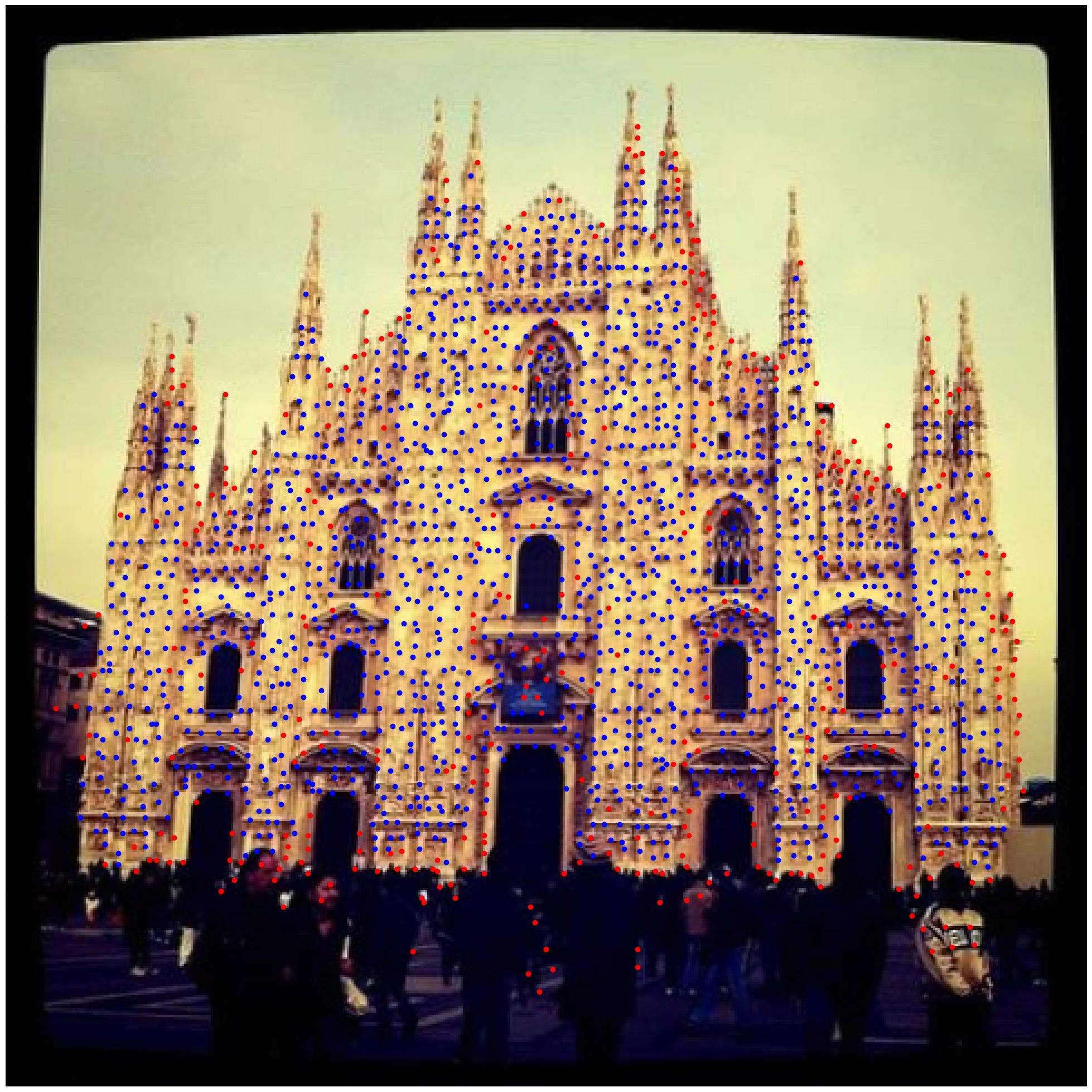}}
\subfloat{\includegraphics[width=0.24\linewidth,height=2.in]{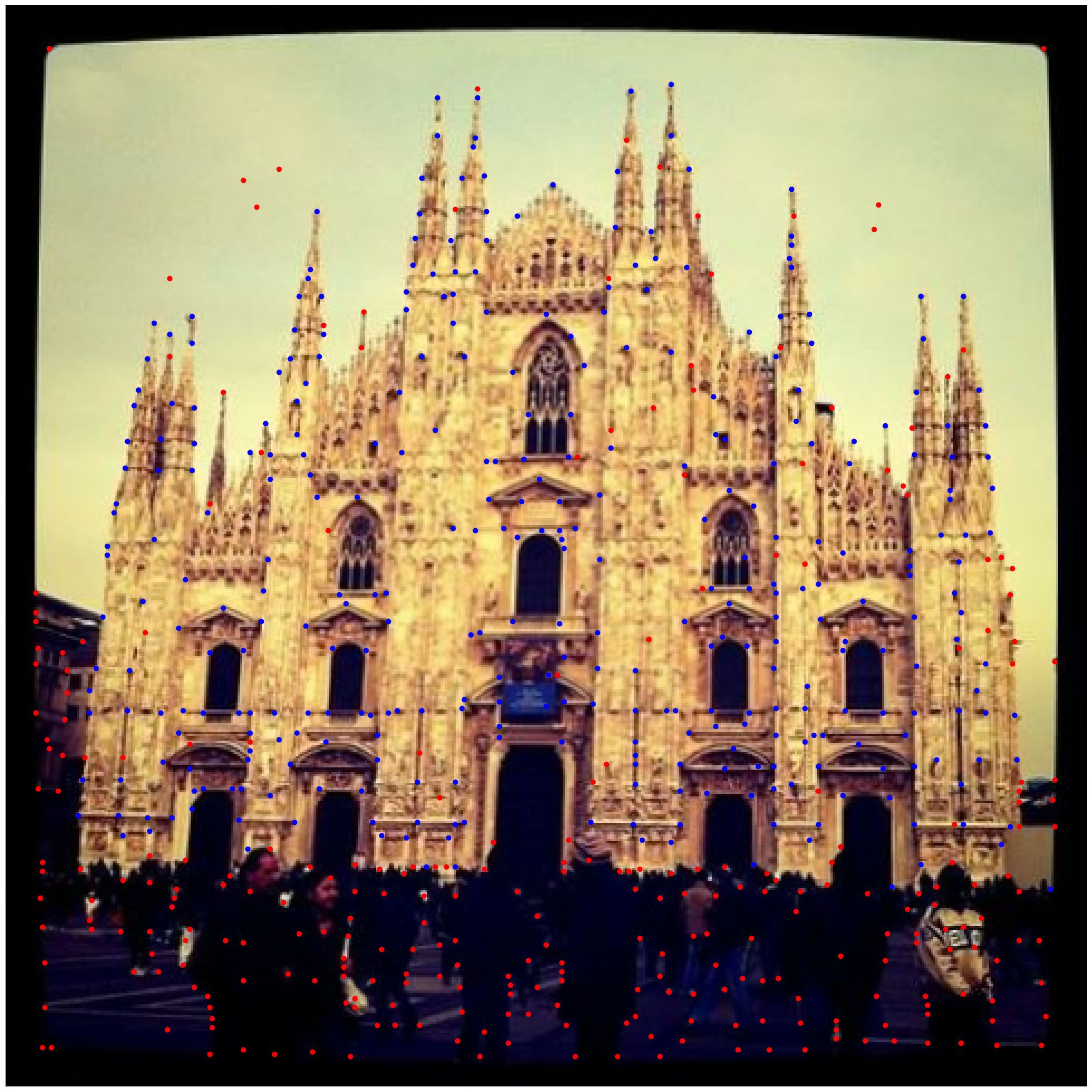}}
\subfloat{\includegraphics[width=0.24\linewidth,height=2.in]{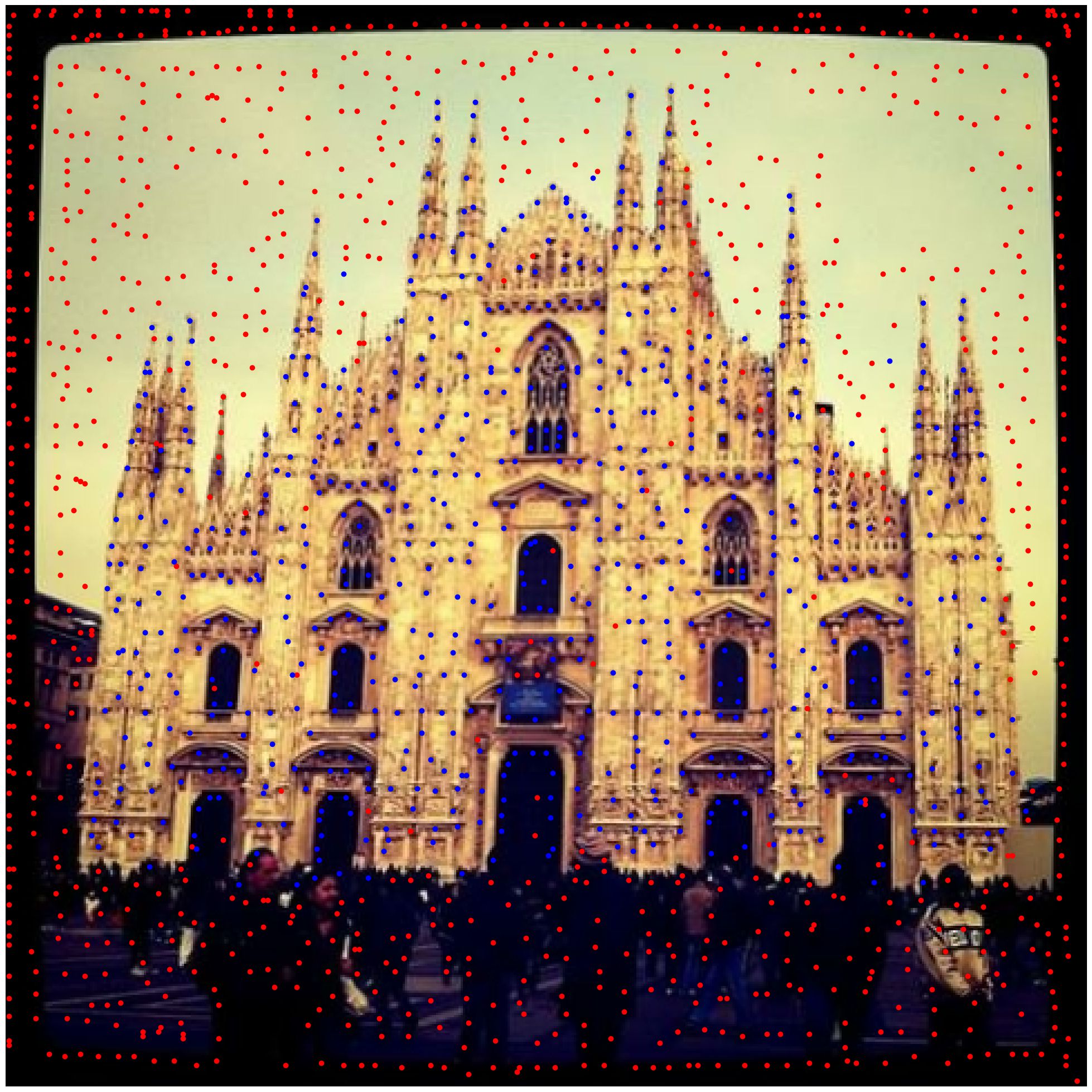}}
\subfloat{\includegraphics[width=0.24\linewidth,height=2.in]{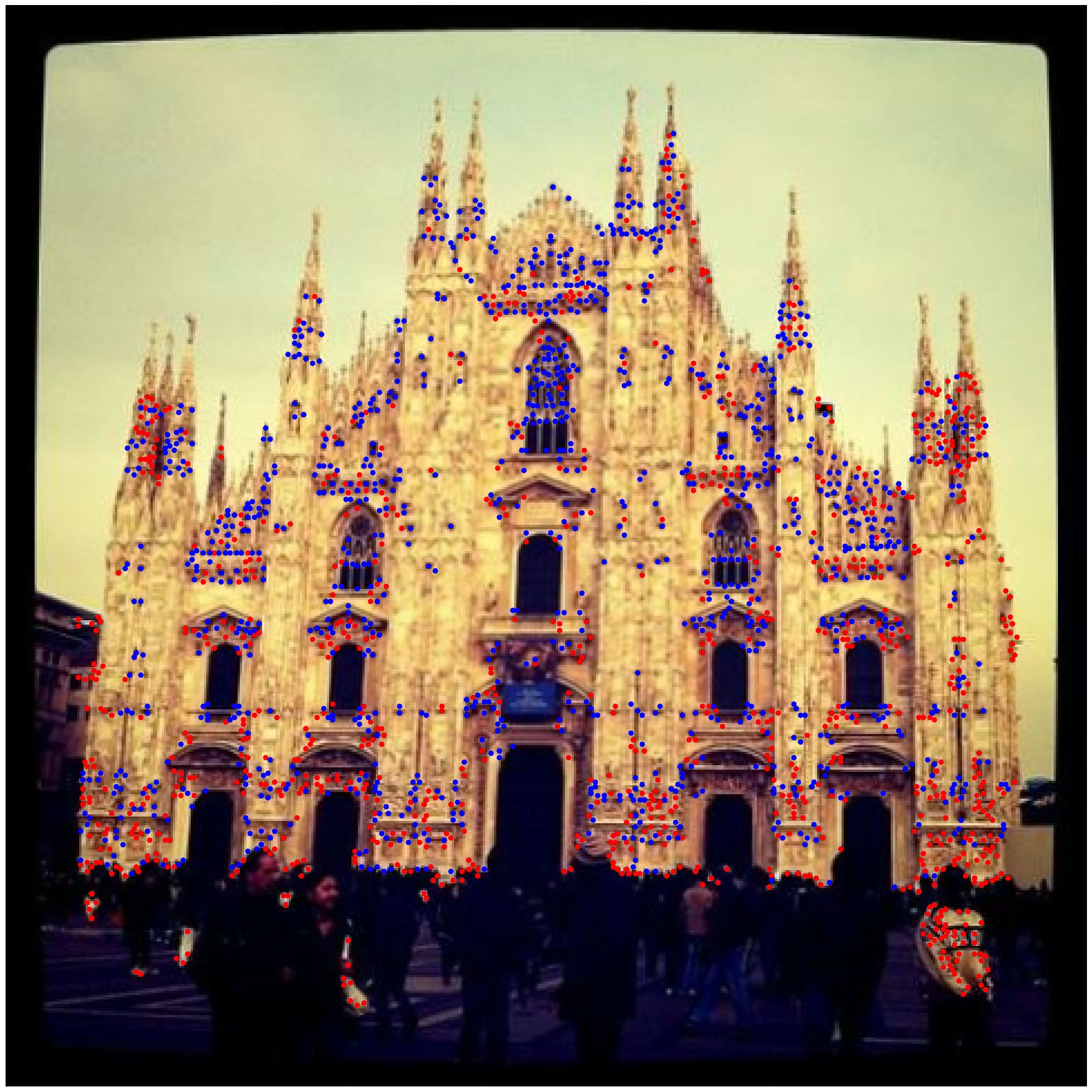}}

\subfloat{\includegraphics[width=0.24\linewidth,height=2.in]{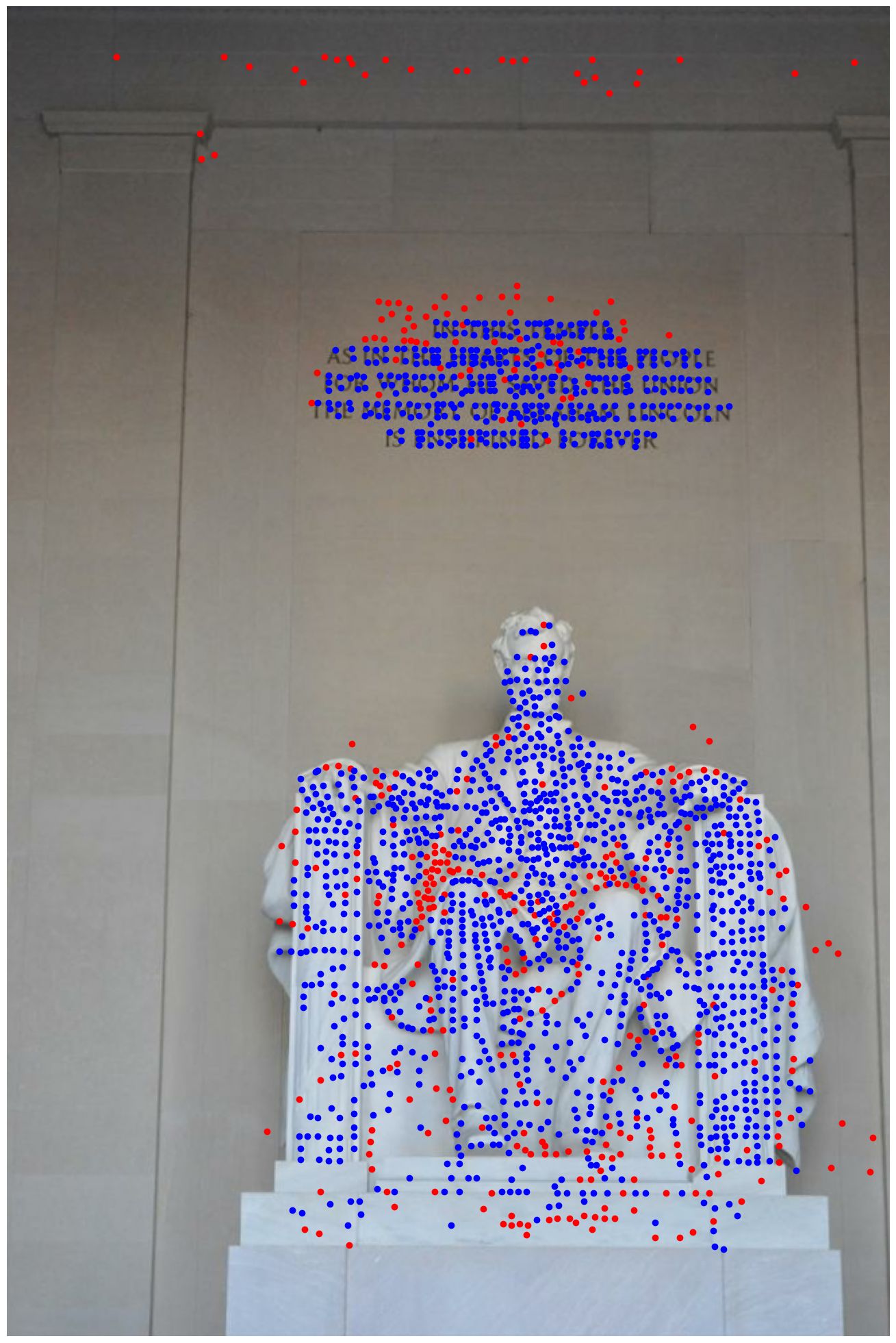}}
\subfloat{\includegraphics[width=0.24\linewidth,height=2.in]{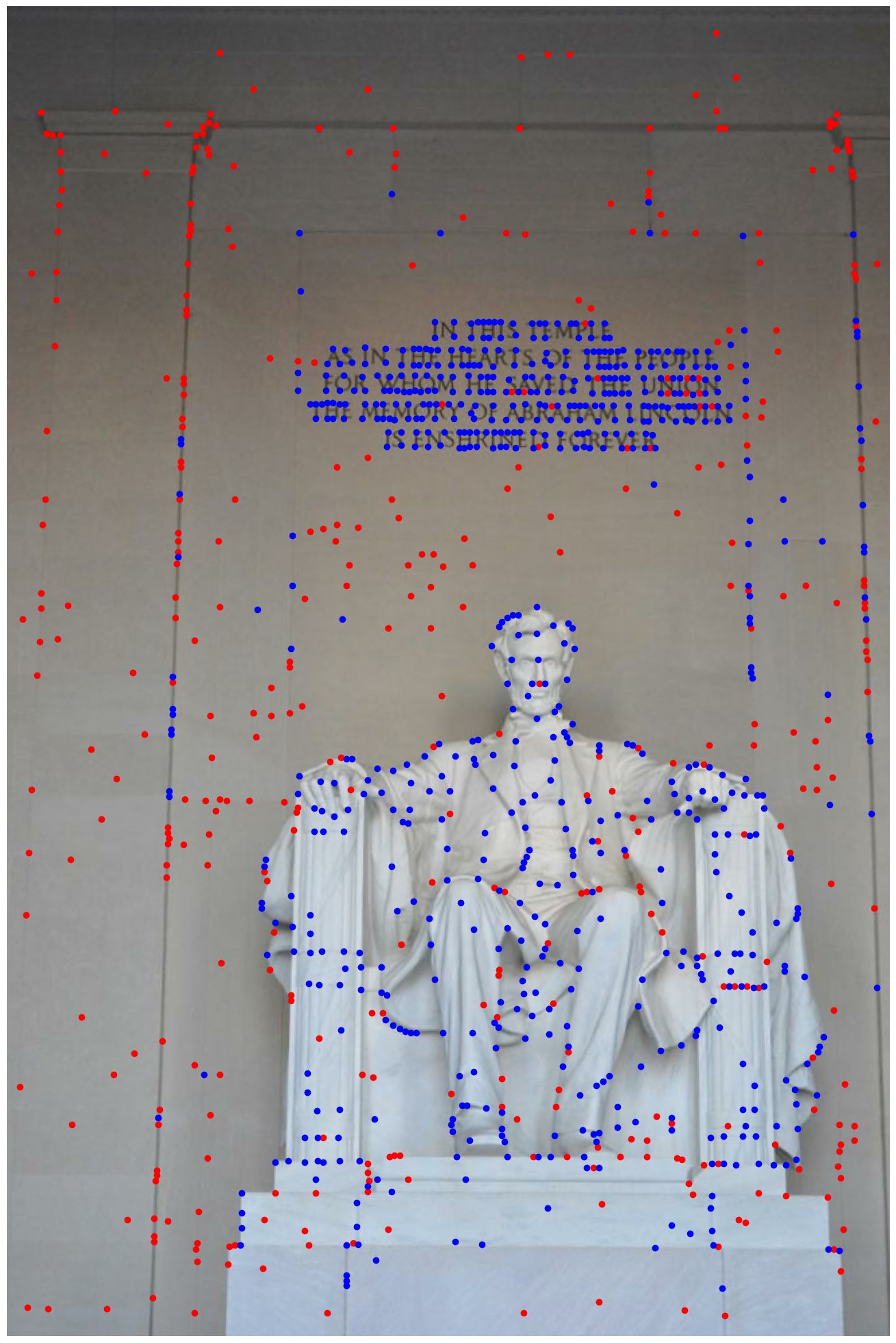}}
\subfloat{\includegraphics[width=0.24\linewidth,height=2.in]{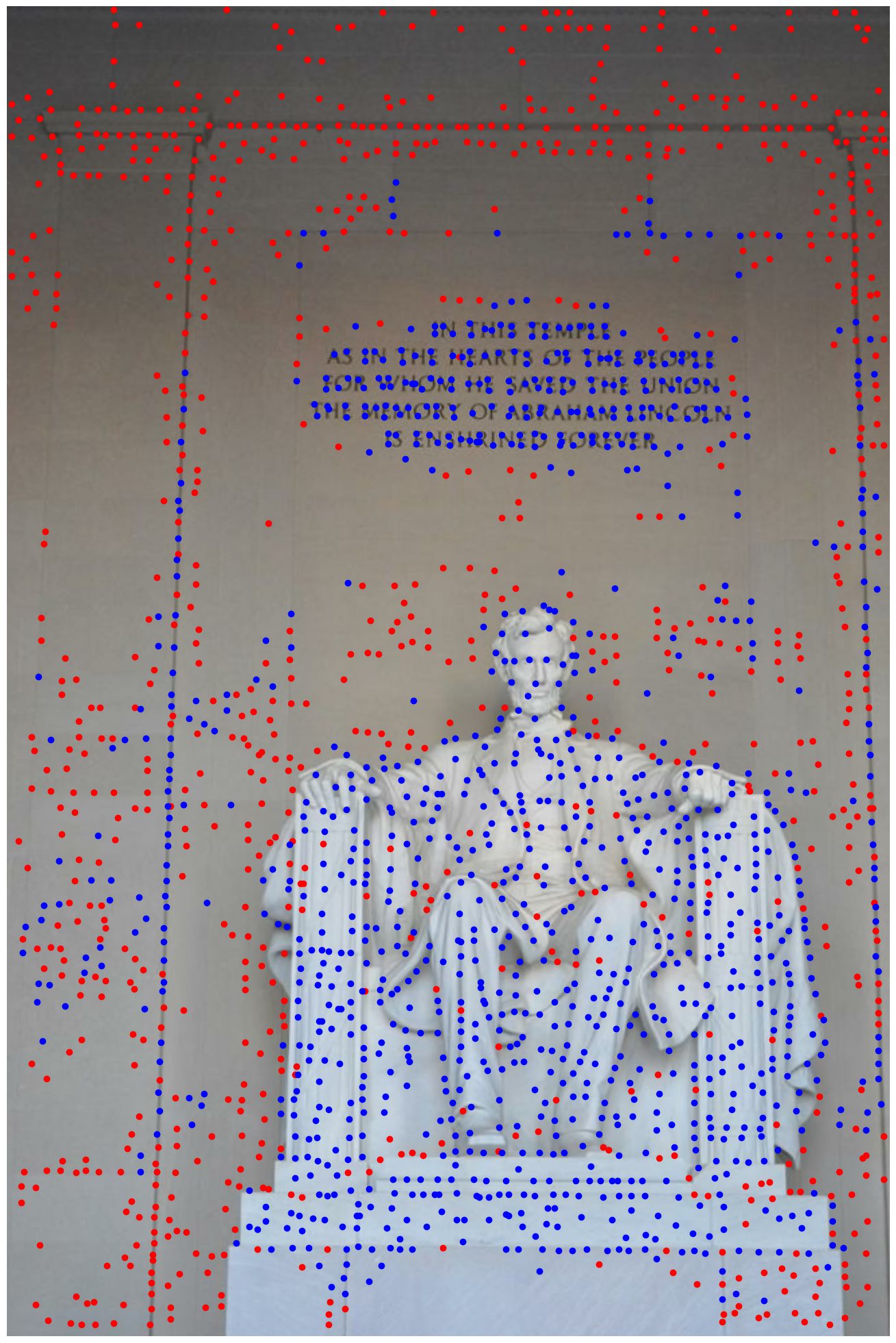}}
\subfloat{\includegraphics[width=0.24\linewidth,height=2.in]{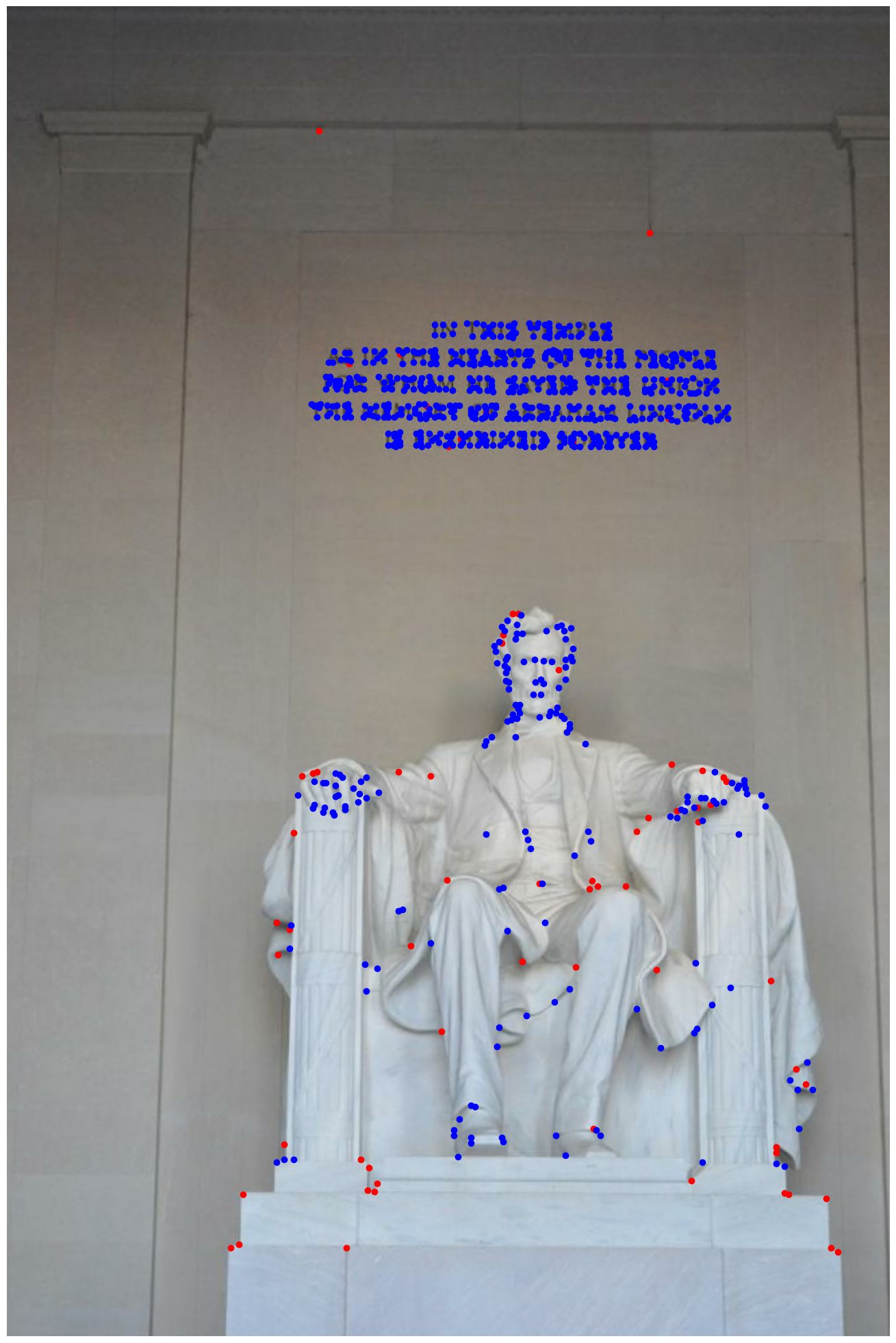}}
 \caption{Results on the IMC2021 multi-view task. All results are obtained using our single LightGlue model fine-tuned for the DISK descriptors. From left to right, keypoints are detected by DISK, SuperPoint, R2D2 and ORB. Keypoints that are part of the 3D reconstruction models are shown in blue, while others are shown in red. The results demonstrate that our model can effectively match keypoints from various detectors. Best viewed in color.
}
 \label{fig:vis_imc_multi}
\end{figure*}

We illustrate the performance of our single fine-tuned model applied to different detectors in \cref{fig:vis_imc_multi} on the multi-view task for IMC2021. All results are obtained using single model with DISK descriptors, while the keypoints are detected by DISK, SuperPoint, R2D2 and ORB. 
Our model effectively matches keypoints from different detectors, while performance differences between features can be attributed to the detectors themselves. For example, the DISK detector predominantly identifies keypoints on buildings and statue, which contributes to its strong performance on tourism photos. SuperPoint detects keypoints at sparser locations, focusing primarily on corners. R2D2 produces a more evenly distributed set of keypoints, which is advantageous for visual localization tasks, as the matches are spread across the image. In contrast, ORB features tend to cluster keypoints, making them less suitable for pose estimation tasks.
\definecolor{Gray}{gray}{0.9}

\subsection{Limitations}
The performance of our model is limited by capacity of descriptor and detector, as we don't fine-tune the weights of local features. 
For now, we need datasets with multi-view posed images such as MegaDepth and MegaScenes \cite{tung2024megascenes} for fine-tuning, which are challenging to collect.

\clearpage
{\small
\bibliographystyle{ieeenat_fullname}
\bibliography{11_references}
}

\end{document}